\documentclass[journal]{IEEEtran}
\usepackage{cite}
\usepackage{amsmath,amssymb,amsfonts}
\usepackage{algorithmic}
\usepackage{graphicx}
\usepackage{textcomp}
\usepackage{booktabs,multicol,multirow}
\usepackage{algorithm}
 \usepackage[colorlinks=true, citecolor=blue, urlcolor=blue, linkcolor=black]{hyperref}
\def\BibTeX{{\rm B\kern-.05em{\sc i\kern-.025em b}\kern-.08em
    T\kern-.1667em\lower.7ex\hbox{E}\kern-.125emX}}
    
\usepackage{color}    
\usepackage[normalem]{ulem}
\usepackage{soul}

\newcommand{\bftab}{\fontseries{b}\selectfont}

\begin{document}
% \history{Date of publication xxxx 00, 0000, date of current version xxxx 00, 0000.}
% \doi{10.1109/ACCESS.2017.DOI}

%%%%%%%%%%%%%%%%%%%%%%%%%%%%%%%%%%%%%%%%%%%%%%%%%%%%%%%%%%%%%%%%%%%%%%%%%%%%%%%%%%

\title{Headline-Guided Extractive Summarization for Thai News Articles}

%%%%%%%%%%%%%%%%%%%%%%%%%%%%%%%%%%%%%%%%%%%%%%%%%%%%%%%%%%%%%%%%%%%%%%%%%%%%%%%%%%

% \author{
% \uppercase{Ukrit Watchareeruetai}\authorrefmark{1},
% \uppercase{Benjaphan Sommana}\authorrefmark{1}, 
% \uppercase{Sanjana Jain}\authorrefmark{1}, 
% \uppercase{Pavit Noinongyao}\authorrefmark{1}, 
% \uppercase{Ankush Ganguly}\authorrefmark{1}, 
% \uppercase{Aubin Samacoits}\authorrefmark{1}, 
% \uppercase{Samuel W.F. Earp}\authorrefmark{1}, 
% \uppercase{and Nakarin Sritrakool}\authorrefmark{2},
% }
% \address[1]{Sertis Vision Lab, 597/5 Sukhumvit Road, Watthana, Bangkok, 10110, Thailand}
% \address[2]{Department of Mathematics and Computer Science, Faculty of Science, Chulalongkorn University, Phayathai Road, Pathum Wan, Bangkok 10330, Thailand}

% \tfootnote{Nakarin Sritrakool has contributed to this work during his internship at Sertis Vision Lab.}

\markboth
% {Watchareeruetai \headeretal: LOTR: Face Landmark Localization Using Localization Transformer}
% {Watchareeruetai \headeretal: LOTR: Face Landmark Localization Using Localization Transformer}
% {Citation information: DOI 10.1109/ACCESS.2022.3149380, IEEE Access}
% {Citation information: DOI 10.1109/ACCESS.2022.3149380, IEEE Access}
% {Watchareeruetai \headeretal: LOTR: Face Landmark Localization Using Localization Transformer, IEEE Access, DOI: 10.1109/ACCESS.2022.3149380.}
% {Watchareeruetai \headeretal: LOTR: Face Landmark Localization Using Localization Transformer IEEE Access, DOI: 10.1109/ACCESS.2022.3149380.}
{Accepted for publication in IEEE Access, 2025 but has not been fully edited. DOI: 10.1109/ACCESS.2025.3538329}
{Accepted for publication in IEEE Access, 2025 but has not been fully edited. DOI: 10.1109/ACCESS.2025.3538329}

\author{\IEEEauthorblockN{
        Pimpitchaya Kositcharoensuk\IEEEauthorrefmark{1},
        Nakarin Sritrakool\IEEEauthorrefmark{2},
        Ploy N. Pratanwanich\IEEEauthorrefmark{1}\IEEEauthorrefmark{3}\\
        }
\IEEEauthorblockA{\IEEEauthorrefmark{1}Department of Mathematics and Computer Science, Faculty of Science, Chulalongkorn University,
Bangkok 10330, Thailand\\}
\IEEEauthorblockA{\IEEEauthorrefmark{2}National Institute of Informatics, Hitotsubashi, Chiyoda, Tokyo 1018430, Japan\\}
\IEEEauthorblockA{\IEEEauthorrefmark{3}Chula Intelligent and Complex Systems Research Unit, Chulalongkorn University, Bangkok 10330, Thailand}
\thanks{Corresponding author: Ploy N. Pratanwanich (e-mail: naruemon.p@chula.ac.th.)}
}

%%%%%%%%%%%%%%%%%%%%%%%%%%%%%%%%%%%%%%%%%%%%%%%%%%%%%%%%%%%%%%%%%%%%%%%%%%%%%%%%%%
%%%%%%%%%%%%%%%%%%%%%%%%%%%%%%%%%%%%%%%%%%%%%%%%%%%%%%%%%%%%%%%%%%%%%%%%%%%%%%%%%%
%%%%%%%%%%%%%%%%%%%%%%%%%%%%%%%%%%%%%%%%%%%%%%%%%%%%%%%%%%%%%%%%%%%%%%%%%%%%%%%%%%
\maketitle

\begin{abstract}
Text summarization is a process of condensing lengthy texts while preserving their essential information. Previous studies have predominantly focused on high-resource languages, while low-resource languages like Thai have received less attention. 
Furthermore, earlier extractive summarization models for Thai texts have primarily relied on the article's body, without considering the headline. 
This omission can result in the exclusion of key sentences from the summary.
To address these limitations, we propose CHIMA, an extractive summarization model that incorporates the contextual information of the headline for Thai news articles. 
Our model utilizes a pre-trained language model to capture complex language semantics and assigns a probability to each sentence to be included in the summary. By leveraging the headline to guide sentence selection, CHIMA enhances the model's ability to recover important sentences and discount irrelevant ones. 
Additionally, we introduce two strategies for aggregating headline-body similarities, simple average and harmonic mean, providing flexibility in sentence selection to accommodate varying writing styles. Experiments on publicly available Thai news datasets demonstrate that CHIMA outperforms baseline models across ROUGE, BLEU, and F1 scores. 
These results highlight the effectiveness of incorporating the headline-body similarities as model guidance.
The results also indicate an enhancement in the model's ability to recall critical sentences, even those scattered throughout the middle or end of the article. 
With this potential, headline-guided extractive summarization offers a promising approach to improve the quality and relevance of summaries for Thai news articles.
\end{abstract}

%%%%%%%%%%%%%%%%%%%%%%%%%%%%%%%%%%%%%%%%%%%%%%%%%%%%%%%%%%%%%%%%%%%%%%%%%%%%%%%%%% 


% \titlepgskip=-15pt

\section{Introduction}
\label{sec:introduction}
In the digital age, the exponential growth of information presents a significant challenge for users in efficiently retrieving crucial information. Text summarization has emerged as a key solution to condense extensive textual data into concise summaries, retaining the core content. To create an effective summary, summarization models must grasp a deep understanding of natural language, extending beyond the literal meaning of individual words and sentences.

Text summarization techniques can be broadly classified into two categories: abstractive and extractive summarization. Abstractive summarization generates a summary that may contain rephrased or entirely new sentences not present in the original text, aiming for a human-like synthesis of the content. Despite its potential to produce a more coherent and contextual summary, the abstractive method often faces challenges such as grammatical errors, hallucinations, and factual inconsistencies \cite{chatgptsum}. These issues pose significant obstacles to their reliability for delivering precise information \cite{chatgptsum}. In contrast, extractive summarization focuses on selecting significant sentences directly from the source text. This approach ensures that the extracted summary remains grammatically correct and faithful to the original content. Although an extractive summary may sometimes lack the flexibility and cohesiveness characteristic of abstractive approaches, it provides a reliable and accurate representation of the source text \cite{bertsum}. Consequently, in this work, we focus on extractive summarization due to its robustness and reliability in maintaining the integrity of the original information.

Although advancements in text summarization have been achieved, most research works have concentrated on high-resource languages, particularly English \cite{summareranker, bertsum, matchsum, chatgptsum, sum_survey}. This focus has led to a significant gap in research for low-resource languages, such as Thai, which remain underexplored. Previous works on Thai text summarization \cite{economic_new, th_mf, rel_work_th2}, have incorporated recurrent neural networks (RNNs) such as long short-term memory (LSTM) \cite{lstm} and gated recurrent unit (GRU) \cite{gru} models, to output summaries. 
% While these networks have demonstrated some success, their performance often suffers from the limited linguistic resources in Thai. 
While these networks have demonstrated some success, their performance often suffers in modeling long-term dependencies effectively. 
These RNNs may struggle to capture the literal meaning of individual words and sentences when the text is long, hindering their generalizability. 
Recently, \cite{thaisum_github} attempted to employ a pre-trained language model such as BERT \cite{bert} for Thai summarization. 
However, relying solely on the article's body may fail to capture critical contextual information, especially when important sentences appear later in the text.
For news articles, headlines often provide a concise and focused outline of the entire article, making them a valuable resource for guiding the selection of key sentences in the article’s body.

{}{To address these limitations, we propose a headline-guided extractive summarization model, called \textbf{CHIMA}. The proposed model integrates the headline as an additional feature for sentence selection in Thai news articles.} 
Our contributions are summarized as follows:
\begin{itemize}
    \item{We propose an extractive text summarization model that incorporates the headline information to recover important sentences that may be missed by the classifier based solely on the article's body in the Thai language.}
    \item {We introduce two aggregation functions for selecting summary sentences: simple average (SA) and harmonic mean (HM). These functions combine the sentence selection scores estimated from the article's body with the headline-body similarity scores, effectively assigning sentence probabilities to be included in the summary.}
    \item{We validate our proposed CHIMA models on publicly available Thai news articles.}
\end{itemize}

The remainder of this paper is structured as follows: Section \ref{sec:rel_work} provides further details on related works. Section \ref{sec:method} introduces the task definition and proposed methodology in details. Section \ref{sec:exp} describes our experimental settings and corresponding results. Section \ref{sec:discuss} offers a discussion of our findings, and the paper concludes in Section \ref{sec:conclusion}.

\section{Related work}\label{sec:rel_work}
A major advancement in natural language processing (NLP) came with the introduction of bidirectional encoder representations from transformers (BERT) \cite{bert}. BERT revolutionized NLP by enabling bidirectional comprehension of contextual word embeddings using the transformer architecture \cite{transformer}. 
BERT utilizes the cloze task and the next sentence prediction to enhance the model's ability to understand the context of a word within the entire sentence.
This breakthrough has improved performance across various NLP domains, including sentiment analysis, semantic textual similarity, and question-answering systems.

{}{Several models leverage pre-trained language models for extractive summarization tasks. These models can be classified into two main categories: single-stage and two-stage. In single-stage models, the model directly outputs the probability of each sentence being included in the summary. For example, PreSumm \mbox{\cite{presumm}} employs a pre-trained BERT model to encode articles and obtain contextualized embeddings for their sentences. It emphasizes both abstractive and extractive summarization. On the other hand, BERTSUM \mbox{\cite{bertsum}} focuses solely on extractive summarization, utilizing three distinct types of summarization layers: a simple classifier, an inter-sentence transformer, and a recurrent neural network.
Although effective, single-stage models like BERTSUM are highly dependent on the prediction scores of individual sentences.
This reliance can lead to overconfidence in selecting certain sentences while overlooking others that may also be significant for summarization. These drawbacks often arise because single-stage models struggle to incorporate external contextual information, such as headlines, which can provide valuable guidance in identifying the most relevant sentences. As a result, the performance of single-stage models may suffer, especially in cases where critical information is distributed across sentences that are not prioritized based solely on prediction scores.

In contrast, two-stage models adopt a more refined approach to summarization by first identifying candidate sentences and then re-ranking them based on additional criteria, such as contextual relevance. This approach enables two-stage models to incorporate auxiliary features, such as headlines or semantic coherence, which help mitigate the limitations of over-reliance on prediction scores observed in the single-stage models. For instance, MATCHSUM \mbox{\cite{matchsum}} introduces a semantic text-matching framework that employs a Siamese-BERT \mbox{\cite{sbert}} architecture. This method achieves superior performance by comparing the semantic similarity between articles and candidate summaries. However, MATCHSUM does not explicitly leverage the information provided in headlines, which typically convey the main idea of an article. Consequently, it may struggle to prioritize sentences that align closely with a headline’s core message, potentially leading to summaries that fail to capture the article’s primary focus.

Recently, large language models (LLMs) have shown promising performance on several NLP tasks. GPTSUM \mbox{\cite{chatgptsum}} explores the application of ChatGPT for summarization, demonstrating the potential of LLMs to generate faithful and coherent summaries. However, \mbox{\cite{chatgptsum}} reported that ChatGPT underperforms supervised fine-tuning methods in extractive summarization tasks.}

In the context of the summarization model for Thai articles, 
\cite{th_mf} is one of the early works that proposed an extractive summarization model via matrix factorization.
It constructs a word-sentence matrix and applies a matrix factorization algorithm such as singular value decomposition (SVD) to retrieve sentence representation.
The summary sentences are selected based on their importance score such as singular value in the SVD.
The work \cite{th_fast} proposed a hybrid summarization model that combines extractive and abstractive approaches. The sentences in the article were first selected using the PageRank algorithm and then passed to mBART \cite{bart} to generate an abstractive summary. However, these models lack optimization during the extractive process, limiting their ability to effectively learn how to select important sentences for the summary.
EconoThai \cite{economic_new} proposed to utilize recurrent neural networks such as LSTM and GRU to output an abstractive summary.
The work \cite{th_key} also proposed to apply an LSTM, which utilizes the keyword as an additional input to generate the abstractive summarization.
Despite the existence of these models for summarizing Thai articles, their performance remains inadequate for practical, real-world applications.

To this end, there has been limited exploration of extractive summarization models for the Thai language that exploit pre-trained language models. Particularly, none of the previous works have been interested in utilizing headlines to recover important sentences that the model might otherwise miss.

\begin{figure*}[t]
\centering
\includegraphics[width = 1.0\textwidth]{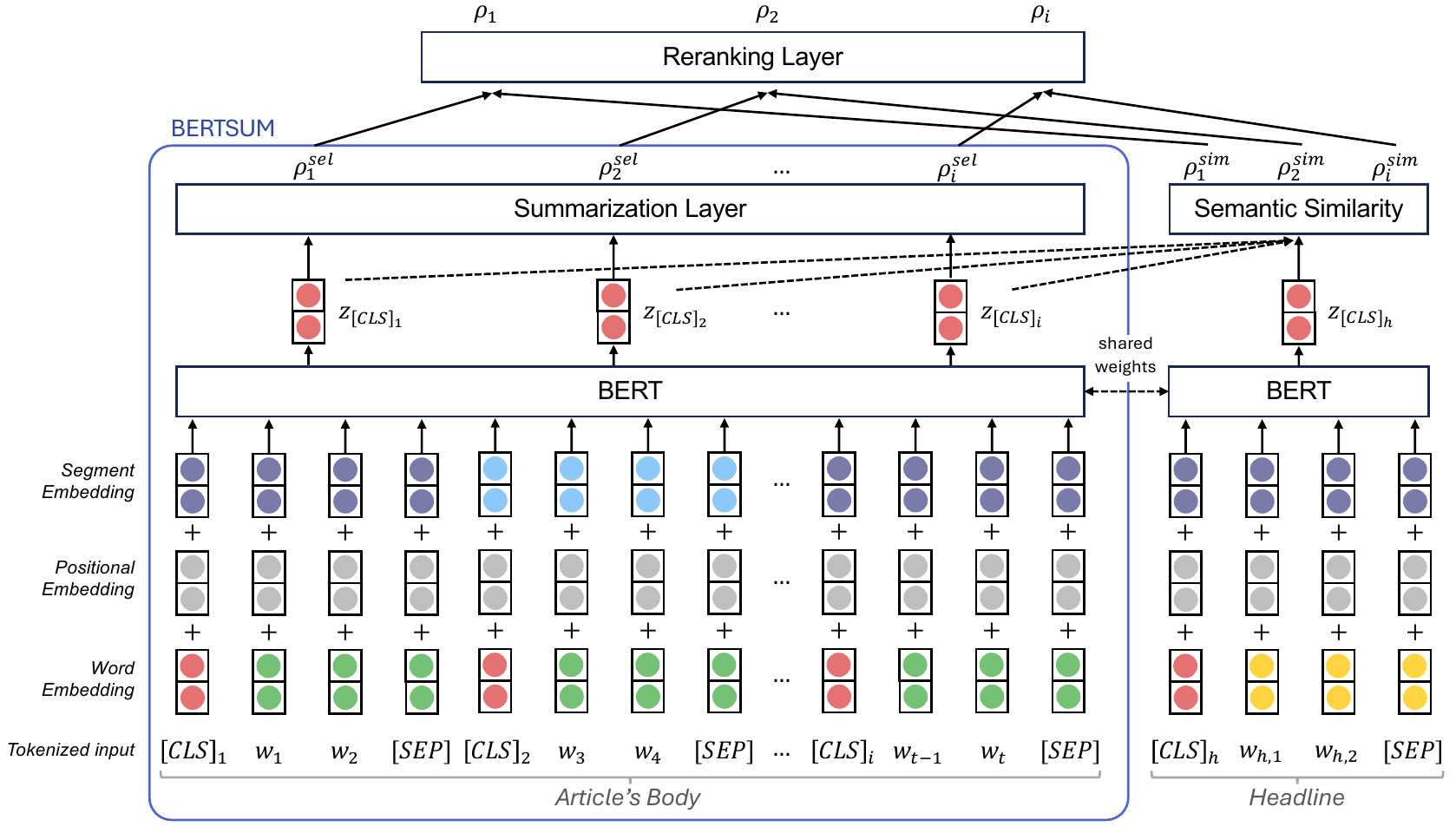}
\caption{Illustration of the proposed CHIMA model.}
\label{fig:model}
\end{figure*}

\section{Methodology}\label{sec:method}

\begin{algorithm}[t!]
\caption{Oracle}
\label{alg:oracle}
\begin{algorithmic}[1]
\renewcommand{\algorithmicrequire}{\textbf{Input:}}
\renewcommand{\algorithmicensure}{\textbf{Output:}}

\REQUIRE {sentence in article's body $S_{body}$, abstractive summary $S_{abs}$, maximum summary sentence $\tau$}
\ENSURE  {Label for Extractive Summarization $Y$}
\\
{$\bigtriangledown$ \textbf{ROUGE-1}}
\hrulefill
\STATE{\textbf{Function }ROUGE-1(predicted unigram $\tilde{C}$, target unigram $C^{abs}$)}\label{alg:start_r1}
\STATE{$overlap \gets \{\textrm{unigram} | \textrm{unigram} \in \tilde{C} \cap C^{abs}$\}}
\STATE{$precision \gets {|overlap|}/{|\tilde{C}|}$}
\STATE{$recall \gets {|overlap|}/{|C^{abs}|}$}
\STATE{$\textrm{ROUGE-1}\gets \frac{2\times precision \times recall} {precision + recal}$}\label{alg:end_r1}
\\
{$\bigtriangledown$ \textbf{ROUGE-2}}
\hrulefill
\STATE{\textbf{Function }ROUGE-2(predicted bigram $\tilde{C}$, target bigram $C^{abs}$)}\label{alg:start_r2}
\STATE{$overlap \gets \{\textrm{bigram} | \textrm{bigram} \in \tilde{C} \cap C^{abs}$\}}
\STATE{$precision \gets {|overlap|}/{|\tilde{C}|}$}
\STATE{$recall \gets {|overlap|}/{|C^{abs}|}$}
\STATE{$\textrm{ROUGE-2}\gets \frac{2\times precision \times recall} {precision + recal}$}\label{alg:end_r2}
\\
{$\bigtriangledown$ \textbf{Compute unigram and bigram}}
\hrulefill
\STATE{$C^{body}_{1} \gets$ \{unigram of words in $S_i | S_i \in S_{body}\}$}\label{alg:start_gram}
\STATE{$C^{abs}_{1} \gets$ \{unigram of words in $S_i | S_i \in S_{abs}\}$}
\STATE{$C^{body}_{2} \gets$ \{bigram of words in $S_i | S_i \in S_{body}\}$}
\STATE{$C^{abs}_{2} \gets$ \{bigram of words in $S_i | S_i \in S_{abs}\}$}\label{alg:end_gram}\\
{$\bigtriangledown$ \textbf{Greedy selection}}
\hrulefill
\STATE{$Y \gets \{\}$}
\STATE{$rouge_{max} \gets 0$}
\WHILE{$|Y| < \tau$}
    \STATE{$isSelected \gets False$}
    \FOR{$i=1$ to $|S_{body}|$} \label{alg:start_loop}
        \IF{$i \notin Y$}
        \STATE{$\tilde{Y} \gets  Y \cup i $}
        \STATE{$\tilde{C}_{1} \gets \bigcup_{k \in \tilde{Y}} C^{body}_{1}[k]$ }
        \STATE{$\tilde{C}_{2} \gets \bigcup_{k \in \tilde{Y}} C^{body}_{2}[k]$ }
        \STATE{$rouge_1 \gets \textrm{ROUGE-1}(\tilde{C}_{1}, C^{abs}_{1})$} \label{alg:r1}
        \STATE{$rouge_2 \gets \textrm{ROUGE-2}(\tilde{C}_{2}, C^{abs}_{2})$} \label{alg:r2}
        \STATE{$rouge_{total} \gets rouge_1 + rouge_2$}\label{alg:start_total}
            \IF{$rouge_{total} > rouge_{max}$}
            \STATE{$rouge_{max} \gets rouge_{total}$}
            \STATE{$isSelected \gets True$}\label{alg:end_total}
            \ENDIF 
        \ENDIF
    \ENDFOR \label{alg:end_loop}
    \IF{$isSelected $}
    \STATE{\textbf{break}}
    \ENDIF
    \STATE{$Y \gets \tilde{Y}$}
    
\ENDWHILE
\RETURN{$\{y_k|y_k \in Y\},$ where $ y_1 < y_2 < \cdots < y_{|Y|}$}\label{alg:return}
\end{algorithmic}
\end{algorithm}

\subsection{Task definition}
An extractive summarization can be formulated as a binary classification task.  
Each article's body contains a sequence of sentences $S_{body} = [S_{1}, S_{2}, \ldots, S_{i}]$, where $S_{i}$ represents the $i^{th}$ sentence in the article's body. 
The task involves assigning a probability $\textrm{y}_{i} \in \left[0, 1\right]$ to each $\textrm{S}_{i}$, indicating the possibility of being included in the predicted summary. 
Specifically, a probability of $\textrm{y}_i = 0$ denotes that the sentence should not be included in the predicted summary, whereas $\textrm{y}_i = 1$ signifies that should be included.

\subsection{Label generation by Oracle}
As explained in the task definition, extractive summarization is formalized as a binary classification task. 
However, the dataset used in this paper only provides the summary for abstractive summarization, limiting to directly adopt the provided summary for extractive summarization. 
To overcome label scarcity, we follow \cite{summarunner} by employing the Oracle, a greedy algorithm that selects sentences from the article's body to maximize the ROUGE score based on the provided abstractive summarization.

The details of the ROUGE score computation and the Oracle are explained in Algorithm \ref{alg:oracle}. 
First, the Oracle computes the unigram and bigram of words in every sentence of the article's body ($S_{body}$) and the abstractive summary ($S_{abs}$), as shown in Line \ref{alg:start_gram}-\ref{alg:end_gram}.
Then, the Oracle iteratively selects sentences in the article's body until it reaches a pre-defined maximum number of sentences ($\tau$).
Particularly, the Oracle goes through each sentence $s$ in $S_{body}$ and includes it to the label set ($Y$) if the ROUGE score increases when including $s$ (Line \ref{alg:start_loop}-\ref{alg:end_loop}).

During the ROUGE score calculation stage, the Oracle computes ROUGE-1 (Line \ref{alg:start_r1}-\ref{alg:end_r1}), ROUGE-2 (Line \ref{alg:start_r2}-\ref{alg:end_r2}), and $rouge_{total}$ (Line \ref{alg:start_total}-\ref{alg:end_total}) as a combination of ROUGE-1 and ROUGE-2, which serves as the criteria for including a sentence $s$ into the $Y$.
% The algorithms describe computation of ROUGE-1 and ROUGE-2 are shown in Line \ref{alg:start_r1}-\ref{alg:end_r1} and Line \ref{alg:start_r2}-\ref{alg:end_r2}, respectively.
Finally, the Oracle returns the $Y$ with the indices of selected sentences sorted in ascending order (Line \ref{alg:return}).
We will further use the returned label  $Y$ for training extractive summarization models.

\subsection{CHIMA}
Relying solely on the article's body can be problematic since the model may overlook important sentences. 
To address this issue, we propose an extractive summarization model called CHIMA. 
In particular, we leverage the headline, which contains key topics, to promote those summary sentence candidates that are semantically related to be selected for summarization.
As illustrated in Figure \ref{fig:model}, CHIMA consists of four main parts: an embedding layer, a BERT layer, a summarization layer, and a reranking layer.
In particular, the embedding layer transforms words in each sentence $S_i$ to their corresponding embeddings.
The BERT layer is a stack of Transformer encoder layers to incorporate contextual information of words in the article's body \cite{bert} \cite{transformer}.
The summarization layer computes the selection score as a probability of each sentence to be selected in the extractive summarization \cite{bertsum}.
Lastly, the reranking layer leverages the headline to reprioritize the sentence candidates from the summarization layer.

\subsubsection{Embedding layer}
To prepare sentences in $S_{body}$ for a summarization task, we insert a special token $[CLS]_i$ at the beginning of each sentence $S_i$ to be a sentence representation. 
We also append a special token $[SEP]$ at the end of each sentence to indicate the sentence boundary.
Then, this layer retrieves each word $w_t$ in $S_i$ to output the corresponding learnable word embedding $\mathbf{e}_t \in \mathbb{R}^d$.
Next, this layer outputs the learnable positional embedding $\mathbf{p}_t \in \mathbb{R}^d$ based on the position of each word ($t$) in the article's body.
To indicate words within the same sentence, the learnable segment embedding ($\mathbf{b}$) is added to each word based on the index of sentence ($i$) in the article's body.
In this layer, the $\mathbf{b}_1 \in \mathbb{R}^d $ will be utilized for sentence with $i$ as odd number, while $\mathbf{b}_2 \in \mathbb{R}^d $ will be utilized when $i$ is an even number:
\begin{align}
\mathbf{b}_i &= \left\{\begin{matrix}
\mathbf{b}_1, \text{if $i$ is odd}\\
\;\mathbf{b}_2, \text{if $i$ is even}
\end{matrix}\right.
\end{align}
Finally, this layer incorporates information from every embedding by adding them together to output latent embedding $\mathbf{z}$:
\begin{align}
\mathbf{z}_t &= \mathbf{e}_t + \mathbf{p}_t + \mathbf{b}_i.
\end{align}

\subsubsection{BERT layer}
Given the latent embeddings $\mathbf{z}$ from the previous layer, a BERT layer aims to incorporate contextual information for each word based on a stacked $l-$layers of transformer encoders.
Each encoder layer consists of multi-head self-attention (MHSA) and positional-wise feed-forward network (PFFN).
To leverage information from the lower layers and stabilize the training, the residual connection\cite{residual} and layer normalization (LN)\cite{layer_norm} are applied after MHSA and PFFN.
The goal of MHSA is to compute contextualized embeddings based on different $m$ relations of words within the article's body.
While the PFFN applies the non-linear transformation to each word to model complex structures of language.

To utilize the latent embedding $\mathbf{z}_t$ across different $m$ heads in MHSA, the latent embedding $\mathbf{z}_t$ is split into $m$ parts ($\mathbf{z}^{m}_t \in \mathbb{R}^{d/m}$), with each part corresponding to a different head.
Then, MHSA projects each latent embedding $\mathbf{z}^{m}_t$ by three matrices to output query ($\mathbf{q}^m_{t}$), key ($\mathbf{k}^m_{t}$), and value ($\mathbf{v}^m_{t}$):
\begin{align}
\mathbf{q}^m_{t} = \mathbf{W}^q_m \mathbf{z}_t,
\mathbf{k}^m_{t} = \mathbf{W}^k_m \mathbf{z}_t,
\mathbf{v}^m_{t} = \mathbf{W}^v_m \mathbf{z}_t,
\end{align}
where $\mathbf{W}^q_m,\mathbf{W}^k_m,\mathbf{W}^v_m \in \mathbb{R}^{\frac{d}{m} \times \frac{d}{m}}$ are learnable projection matrices for query, key, and value, respectively.
The MHSA computes the contextualized word embedding based on the relationship between words by a scaled-dot product:
\begin{align}
\mathbf{X}^m = \mathrm{Softmax}(\frac{\mathbf{Q}^m {\mathbf{K}^m}^\top}{\sqrt{d/m}})\mathbf{V}^m,
\end{align}
where $\mathbf{Q}^m$, $\mathbf{K}^m$, and $\mathbf{V}^m$ represent the stacked queries, keys, and values, respectively. $\mathbf{X}^m$ denotes the stacked latent word embeddings.
To combine contextualized embedding from $m$ heads, MHSA concatenates latent embeddings from each head and applies a linear transformation to integrate embeddings ($\mathbf{X}^1,...,\mathbf{X}^m$) into a single embedding:
\begin{align}
\mathbf{X} = \mathbf{W}_0^l[\mathbf{X}^1;\mathbf{X}^2;...;\mathbf{X}^m],
\end{align}
where $\mathbf{W}^l_0 \in \mathbb{R}^{d\times d}$ is a learnable projection matrix.
Then, the residual connection and layer normalization is applied to each $\mathbf{z}_t \in \mathbf{Z}$ as follows:
\begin{align}
\mathbf{z}_t = \mathrm{LN}(\mathbf{x}_t + \mathbf{z}_t).
\end{align}
To introduce a non-linear transformation to the latent representation, the PFFN is applied to each $\mathbf{z}_t$ along with residual connection and LN as follows:
\begin{align}
\mathbf{z}_t &= \mathrm{LN}(\mathrm{PFFN}(\mathbf{z}_t) + \mathbf{z}_t),\\
\mathrm{PFFN}(\mathbf{z}_t) &= \mathrm{ReLU}(\mathbf{W}^l_2(\mathbf{W}^l_1 \mathbf{z}_t + \mathbf{b}^l_1)+\mathbf{b}^l_2),
\end{align}
where $\mathbf{W}^l_1 \in \mathbb{R}^{d \times d}$ and $\mathbf{W}^l_2 \in \mathbb{R}^{d \times d}$ are learnable projection matrices. $\mathbf{b}^l_1 \in \mathbb{R}^{d}$ and $\mathbf{b}^l_2 \in \mathbb{R}^{d}$ are learnable biases.

\subsubsection{Summarization layer}
In this layer, we compute the selection score ($\rho^{sel}_i$) as the probability of each sentence being selected as part of the summary based solely on its semantic meaning.
We represent each sentence $S_i$ by latent $[CLS]_i$ embedding ($\mathbf{z}_{[CLS]_i}$).
We pass each $\mathbf{z}_{[CLS]_i}$ to a feed-forward network with a sigmoid activation function to output the selection score as in \cite{bertsum}:
\begin{align}
\rho^{sel}_i = \sigma(\mathbf{W}_o\mathbf{z}_{[CLS]_i} + \mathbf{b}_o),
\end{align}
where $\mathbf{W}_o \in \mathbb{R}^{1\times d}$ and $\mathbf{b}_o\in\mathbb{R}$ are learnable projection matrix and bias.
We adopt Binary Cross-Entropy loss to optimize the embedding and BERT layer, which measures the correctness of the predictions against the ground truth labels ($Y$) from the Oracle:
\begin{align}
\mathcal{L}(\rho^{sel}_i, y_i) = y_i\mathrm{log}(\rho^{sel}_i) + (1-y_i)\mathrm{log}(1-\rho^{sel}_i).\label{eq:loss}
\end{align}

\subsubsection{Headline-guided reranking layer}
In principle, the headline serves to capture the key points from the article's body, making it essential for summarization. 
Rather than relying solely on the article's body, we propose to leverage the headline information to reprioritize the sentence candidates obtained from the summarization layer. 
To leverage the headline, we insert the special tokens $[CLS]_h$ and $[SEP]$ in the first and last positions in the headline, respectively.
Similar to the article's body, we compute the latent embeddings of the headline ($h$) by passing words in the headline to the optimized embedding and BERT layers:
\begin{align}
\mathbf{Z}_h &= \mathbf{E}_h+\mathbf{P}_h+\mathbf{B}_h,\\
\mathbf{Z}_h &= \mathrm{BERT}(\mathbf{Z}_h),
\end{align}
where $\mathbf{E}_h,\mathbf{P}_H,\mathbf{B}_h$ represent the stacked word embeddings, positional embeddings, and segment embeddings of each word in the headline $h$.
In this work, we first trained the embedding and BERT layers, then applied the reranking layer. We also conducted an experiment where the model was trained in an end-to-end fashion, but this approach yielded worse performance compared to applying the reranking layer separately.

Then, we leverage the headline by selecting sentences from the article's body that have similar semantic meaning to the headline.
Particularly, we compute cosine similarity as a similarity score ($\rho_i^{sim}$) between the headline embedding $\mathbf{z}_{[CLS]_h}$ and each sentence in the article's body ($S_i$):
\begin{align}
\rho_i^{sim} &= \frac{\mathbf{z}_{[CLS]_h} \cdot \mathbf{z}_{[CLS]_i}^{\top}}{\left \| \mathbf{z}_{[CLS]_h} \right \|\left \| \mathbf{z}_{[CLS]_i} \right \|}.
\end{align}
To this end, there are two scores for including sentences in the predicted summary, i.e., selection score ($\rho_i^{sel}$) and similarity score ($\rho_i^{sim}$). 
We propose two strategies to combine both scores as summarization probability ($\rho$) based on aggregation functions: a simple average (CHIMA-SA) and a harmonic mean (CHIMA-HM).
These functions are described as follows:
\begin{align}
    \rho_i &= \mathrm{SA}(\rho_i^{sel},\rho_i^{sim}) = \frac{\rho_i^{sel}+\rho_i^{sim}}{2},\\
\rho_i &= \mathrm{HM}(\rho_i^{sel},\rho_i^{sim}) = \frac{2\times\rho_i^{sel}\times\rho_i^{sim}}{\rho_i^{sel}+\rho_i^{sim}}.
\end{align}
For CHIMA-SA, we assume that the selection score and the similarity score contribute equally to the summarization probability. In contrast, for CHIMA-HM, we assume the consistency between both scores is essentially required in computing the summarization probability.
To output the predicted extractive summary, we select sentences $S_i$ with its summarization probability ($\rho_i$) higher or equal to 0.5:
\begin{align}
    \hat{Y} = \{S_i\mid\rho_i \geq 0.5\}.
\end{align}

\begin{figure}[t]
\centering
\includegraphics[width = 0.48\textwidth]{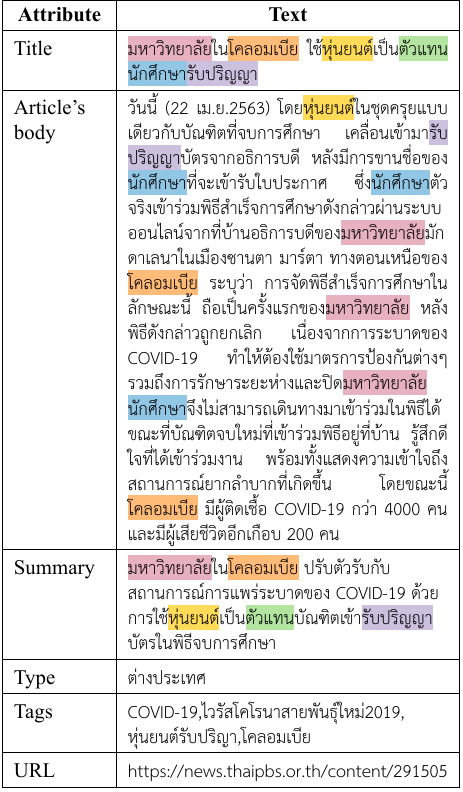}
\caption{A sample from the ThaiSum dataset.}
\label{fig:fig1_dataset_example}
\end{figure}

\begin{figure*}[t]
\centering
\includegraphics[width = 1\textwidth]{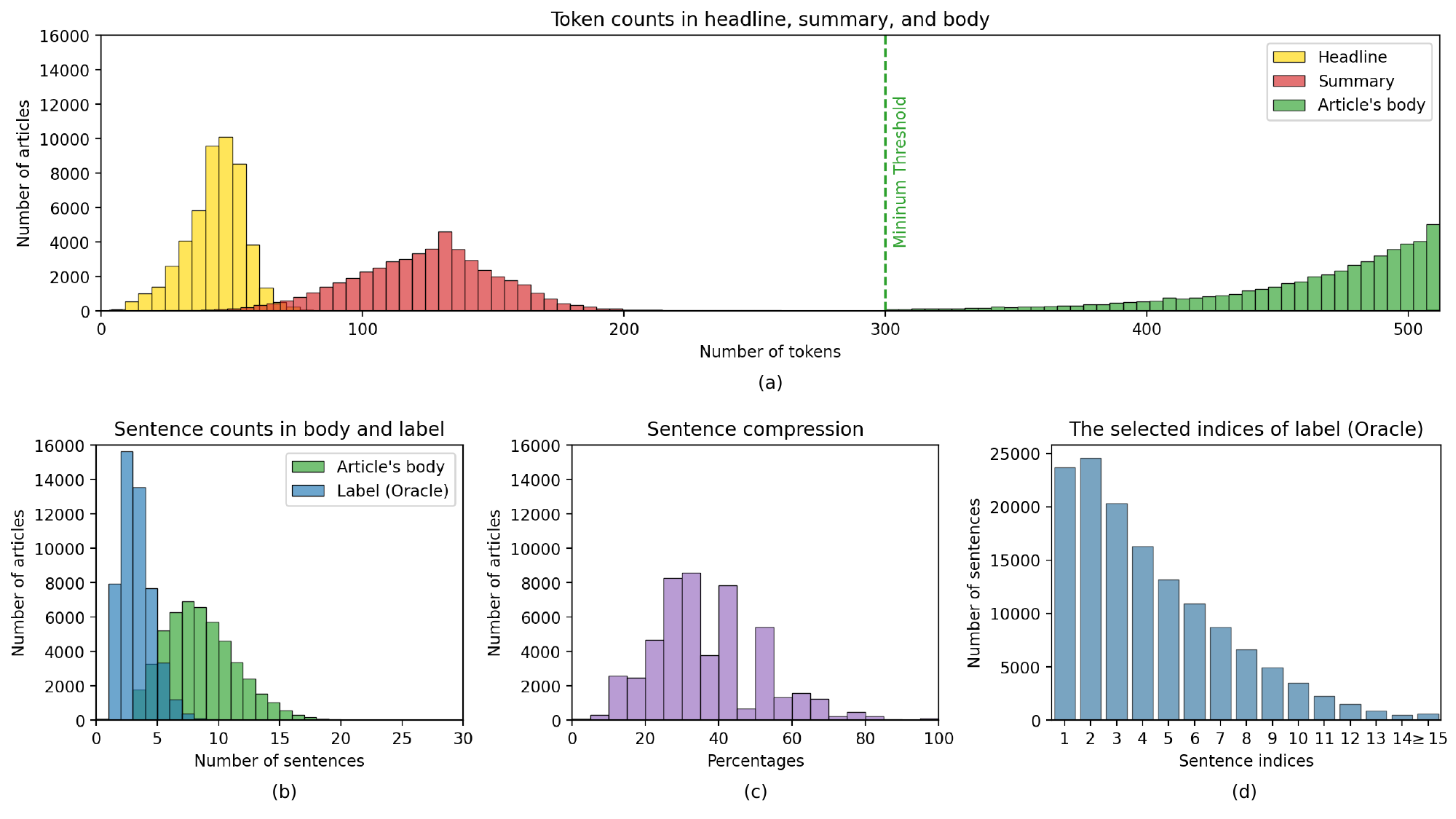}
\caption{Statistics of the ThaiSum dataset. Distributions of the number of tokens in the headline, summary, and article's body parts (a). Distribution of the number of sentences (b), the body-summary compression percentages (c), and the sentence indices of all summary labels from Oracle (d).}
\label{fig:fig3_dataset_analysis}
\end{figure*}

\section{Experiments}\label{sec:exp}

\subsection{Dataset}
We performed our experiments of text summarization tasks on the ThaiSum{\footnote{https://huggingface.co/datasets/thaisum}} benchmark dataset.
This dataset contains Thai news articles and their metadata crawled from various online news sources. This dataset includes Thairath{\footnote{https://www.thairath.co.th}}, Prachatai{\footnote{https://prachatai.com}}, Thai PBS{\footnote{https://www.thaipbs.or.th}}, and The Standard{\footnote{https://thestandard.co}}. Each article consists of six attributes: headline, article’s body, summary, type, tags, and URL, as shown in Figure \ref{fig:fig1_dataset_example}. In this work, we used three attributes from the provided dataset: headline, article’s body, and summary. For computational feasibility, we limited our training set to 50,000 samples, with two additional sets of 5,000 samples, each of which was held out for validation and testing. Only those articles containing between 3 and 30 sentences were included, where each sentence was constrained to contain at least 300 tokens but no more than 512 tokens.

The data preprocessing pipeline consists of several stages. First, we cleaned the data by eliminating articles with missing values and duplicates, keeping only the initial instance of each duplicate entry. Regular expressions were employed to strip out HTML tags, non-printable characters, and specific Unicode ranges. Backslashes were replaced with spaces. We performed text normalization to maintain its consistency by standardizing zero-width spaces and duplicate diacritics (spaces, vowels, signs, and tone marks) using PyThaiNLP \cite{pythainlp}. Finally, we applied sentence and word tokenization using the PyThaiNLP library and a BERT tokenizer to transform the raw text into smaller units known as tokens.

The distributions of token counts across headlines, summaries, and articles' bodies reveal distinct patterns in content length, as shown in Figure \ref{fig:fig3_dataset_analysis}(a). 
Headlines demonstrate a narrow distribution, peaking at around 50 tokens, which reflects their succinct nature. 
Summaries, with a broader distribution centered around 130 tokens, capture a more detailed overview. In contrast, the articles' bodies exhibit a wide-ranging distribution with a peak near 500 tokens, indicating their comprehensive and expansive scope. These distributions align with expected structures: headlines are concise, summaries provide a more detailed overview, and articles' bodies contain extensive content. 
{}{The distinct peaks and variability in token distributions across headlines, summaries, and article bodies illustrate the dataset-specific characteristics. As the number of articles containing fewer than 300 tokens in the body is notably low, we set a minimum threshold of 300 tokens as shown in Figure \mbox{\ref{fig:fig3_dataset_analysis}(a)}}.

Following a similar trend as observed in Figure \ref{fig:fig3_dataset_analysis}(a), the number of sentences in articles' bodies tends to be higher. At the same time, the corresponding labels (Oracle) show a smaller distribution concentrated in fewer sentence lengths (see Figure \ref{fig:fig3_dataset_analysis}(b)). Additionally, the proportion of sentences compressed from the articles' bodies into the summaries is centered around 30-50\%, as shown in Figure \ref{fig:fig3_dataset_analysis}(c).

Figure \ref{fig:fig3_dataset_analysis}(d) illustrates the frequency of summary sentences selected as the extractive labels at different indices within articles. The x-axis represents the sentence indices, and the y-axis represents the number of sentences. The $1^{\textrm{st}}$ index refers to the first sentence of the article's body, the $2^{\textrm{nd}}$ index refers to the second sentence of the article's body, and so on. Figure \ref{fig:fig3_dataset_analysis}(d) implies that the first few sentence indices i.e., $1^{\textrm{st}}$, $2^{\textrm{nd}}$, and $3^{\textrm{rd}}$, are the most commonly selected and appear to decay exponentially.

\begin{table*}[t!]
\centering
\caption{Model evaluation on human-written abstractive and Oracle-extractive summarization across all news sources using ROUGE and BLEU scores.}
\resizebox{1.0\textwidth}{!}{ % control the size of the table
\begin{tabular}{llcccccccc}
\toprule
 &  & \multicolumn{4}{c}{Abstractive Summarization} & \multicolumn{4}{c}{Extractive Summarization} \\
\cmidrule(lr){3-6}
\cmidrule(l){7-10}
 &  & \multicolumn{1}{c}{ROUGE-1} & \multicolumn{1}{c}{ROUGE-2} & \multicolumn{1}{c}{ROUGE-L} & \multicolumn{1}{c}{BLEU} & \multicolumn{1}{c}{ROUGE-1} & \multicolumn{1}{c}{ROUGE-2} & \multicolumn{1}{c}{ROUGE-L} & \multicolumn{1}{c}{BLEU} \\
 \midrule
\multirow{10}{*}{\rotatebox[origin=c]{90}{\begin{tabular}{c} All {}{news sources}\\{}{(5,000 articles)} \end{tabular}}} 
% \multirow{10}{*}{\rotatebox[origin=c]{90}{All {}{(5,000 articles)}}}  
 & Oracle & 0.813 & 0.774 & 0.776 & 0.684 & 1.000 & 1.000 & 1.000 & 1.000 \\
 & Lead-2 & 0.651 & 0.585 & 0.596 & 0.443 & 0.717 & 0.673 & 0.681 & 0.545 \\
 & Lead-3 & 0.662 & 0.605 & 0.614 & 0.479 & 0.744 & 0.709 & 0.712 & 0.596 \\
 & Lead-4 & 0.618 & 0.571 & 0.578 & 0.441 & 0.705 & 0.679 & 0.681 & 0.558 \\
 & HL & 0.371 & 0.228 & 0.277 & 0.079 & 0.331 & 0.196 & 0.241 & 0.058 \\
 \cmidrule{2-10}
 & HL-COS & 0.580 & 0.549 & 0.553 & 0.424 & 0.675 & 0.669 & 0.671 & 0.560 \\
 & BERTSUM & 0.718 & 0.654 & 0.663 & 0.558 & 0.815 & 0.779 & 0.786 & 0.713 \\
 & {}{GPTSUM} & {}{0.513} & {}{0.419} & {}{0.422} & {}{0.303} & {}{0.595} & {}{0.524} & {}{0.528} & {}{0.408} \\
 % & GPTSUM & 0.513 & 0.419 & 0.422 & 0.303 & 0.595 & 0.524 & 0.528 & 0.408 \\
 \cmidrule{2-10}
 & CHIMA-HM & \bftab{0.732} & 0.669 & 0.677 & 0.573 & 0.833 & 0.797 & 0.803 & 0.735 \\
 & CHIMA-SA & 0.731 & \bftab{0.671} & \bftab{0.679} & \bftab{0.574} & \bftab{0.834} & \bftab{0.800} & \bftab{0.805} & \bftab{0.738}\\
 \bottomrule
\end{tabular}
} % control the size of the table
\label{tab:result_rouge_bleu_all}
\end{table*}

\subsection{Evaluation metrics and baselines}
\subsubsection{Evaluation metrics}

We utilized ROUGE, BLEU, precision, recall, and F1 scores as our evaluation metrics to measure the performance of all text summarization models in the experiments. ROUGE and BLEU scores are used to determine how well the extracted sentences cover the ground truth summary. Precision, recall, and F1 scores are employed to evaluate the binary classification tasks. 

In particular, we utilized various variants of ROUGE scores \cite{ROUGE}, including ROUGE-N and ROUGE-L (ROUGE-Longest Common Subsequence). ROUGE scores are metrics that measure the overlap between a predicted summary and a target summary. ROUGE-N focuses on matching $n$-gram units, while ROUGE-L measures the length of the longest common subsequence. These scores range from 0 to 1, where a score of 1 indicates an exact match between the predicted summary and the target summary.

\begin{align}
\textrm{ROUGE-N} = \frac{ \sum_{\textrm{gram}_\textrm{n} \in \textrm{S}} \textrm{Count}_{\textrm{match}}(\textrm{gram}_\textrm{n})}{ \sum_{\textrm{gram}_\textrm{n} \in \textrm{S}} \textrm{Count}(\textrm{gram}_\textrm{n})},
\end{align}
where $n$ represents the n-gram length and $S$ is the target summary. $\textrm{gram}_\textrm{{n}}$ denotes an n-gram sequence, and $\textrm{Count}_{\textrm{match}}(\textrm{gram}_\textrm{{n}})$ represents the maximum number of times an n-gram appears in both the predicted summary and the target summary. The ROUGE scores utilized in this experiment include three variants: ROUGE-1, ROUGE-2, and ROUGE-L. All variations of these ROUGE scores are computed based on the F1 score of the matching n-grams.

The BLEU score incorporates a brevity penalty and n-gram precision terms, computing the precision for n-grams of sizes 1 to 4 in the predicted summary relative to the target summary. It computes the frequency of n-grams in the predicted summary that match those in the target summary, while also considering the length of the predicted summary. BLEU also ranges from 0 to 1, with the same interpretation as the ROUGE score.
% \small
\begin{align}
\textrm{BLEU} = \textrm{min}\left ( 1, \frac{\textrm{predicted-length}}{\textrm{target-length}} \right )\left ( \prod_{i=1}^{\textrm{n}}\textrm{precision}_{i} \right )^{\frac{1}{\textrm{n}}},
\end{align}
where $\textrm{min}\left ( 1,\frac{\textrm{predicted-length}}{\textrm{target-length}} \right ) $ serves as a brevity penalty, addressing the issue that shorter text segments may achieve higher n-gram precision by coincidence. This penalty reduces the score when the predicted summary is much shorter than the target summary. The expression $\left ( \prod_{i=1}^{\textrm{n}}\textrm{precision}_{i} \right )^{\frac{1}{\textrm{n}}}$ represents the geometric mean of n-gram precision, where BLEU typically calculates precision for n-grams of sizes ranging from 1 to 4 by comparing the predicted summary to the target summary.

\begin{table*}[]
\centering
\caption{Model evaluation on human-written abstractive and Oracle-extractive summarization across individual news sources using ROUGE and BLEU scores.}
\resizebox{1.0\textwidth}{!}{ % control the size of the table
\begin{tabular}{llcccccccc}
\toprule
 &  & \multicolumn{4}{c}{Abstractive Summarization} & \multicolumn{4}{c}{Extractive Summarization} \\
 \cmidrule(lr){3-6}
\cmidrule(l){7-10}
 &  & \multicolumn{1}{c}{ROUGE-1} & \multicolumn{1}{c}{ROUGE-2} & \multicolumn{1}{c}{ROUGE-L} & \multicolumn{1}{c}{BLEU} & \multicolumn{1}{c}{ROUGE-1} & \multicolumn{1}{c}{ROUGE-2} & \multicolumn{1}{c}{ROUGE-L} & \multicolumn{1}{c}{BLEU} \\
  \midrule
\multirow{10}{*}{\rotatebox[origin=c]{90}
{\begin{tabular}{c}
     Thairath\\(3,436 articles)
\end{tabular}}} 
 & Oracle & 0.810 & 0.775 & 0.779 & 0.683 & 1.000 & 1.000 & 1.000 & 1.000 \\
 & Lead-2 & 0.641 & 0.569 & 0.581 & 0.430 & 0.706 & 0.656 & 0.665 & 0.527 \\
 & Lead-3 & 0.661 & 0.603 & 0.612 & 0.477 & 0.745 & 0.705 & 0.710 & 0.593 \\
 & Lead-4 & 0.623 & 0.576 & 0.584 & 0.447 & 0.713 & 0.683 & 0.686 & 0.565 \\
 & HL & 0.369 & 0.224 & 0.273 & 0.076 & 0.326 & 0.191 & 0.236 & 0.054 \\
  \cmidrule{2-10}
 & HL-COS & 0.588 & 0.561 & 0.566 & 0.435 & 0.689 & 0.682 & 0.684 & 0.575 \\
 & BERTSUM & 0.720 & 0.657 & 0.667 & 0.564 & 0.819 & 0.781 & 0.788 & 0.717 \\
 & {}{GPTSUM} & {}{0.513} & {}{0.423} & {}{0.427} & {}{0.306} & {}{0.600} & {}{0.528} & {}{0.532} & {}{0.412} \\
 % & GPTSUM & 0.513 & 0.423 & 0.427 & 0.306 & 0.600 & 0.528 & 0.532 & 0.412 \\
  \cmidrule{2-10}
 & CHIMA-HM & \bftab{0.735} & 0.673 & 0.682 & 0.578 & \bftab{0.839} & 0.800 & 0.807 & 0.740 \\
 & CHIMA-SA & \bftab{0.735} & \bftab{0.675} & \bftab{0.684} & \bftab{0.579} & \bftab{0.839} & \bftab{0.802} & \bftab{0.808} & \bftab{0.743} \\
 \midrule 
\multirow{10}{*}{\rotatebox[origin=c]{90}{\begin{tabular}{c}
     The Standard\\(136 articles)
\end{tabular}}} & Oracle & 0.671 & 0.427 & 0.414 & 0.335 & 1.000 & 1.000 & 1.000 & 1.000 \\
 & Lead-2 & 0.446 & 0.194 & 0.231 & 0.105 & 0.555 & 0.408 & 0.423 & 0.269 \\
 & Lead-3 & 0.515 & 0.245 & 0.265 & 0.160 & 0.639 & 0.512 & 0.508 & 0.387 \\
 & Lead-4 & 0.552 & 0.278 & 0.285 & 0.199 & 0.678 & 0.582 & 0.575 & 0.457 \\
 & HL & 0.235 & 0.119 & 0.161 & 0.024 & 0.249 & 0.123 & 0.166 & 0.021 \\
   \cmidrule{2-10}
 & HL-COS & \bftab{0.585} & \bftab{0.326} & \bftab{0.310} & \bftab{0.247} & 0.714 & \bftab{0.710} & \bftab{0.713} & \bftab{0.582} \\
 & BERTSUM & 0.518 & 0.272 & 0.284 & 0.171 & 0.670 & 0.591 & 0.602 & 0.458 \\
 & {}{GPTSUM} & {}{0.574} & {}{0.307} & {}{0.301} & {}{0.232} & {}{0.711} & {}{0.650} & {}{0.654} & {}{0.535} \\
 % & GPTSUM & 0.574 & 0.307 & 0.301 & 0.232 & 0.711 & 0.650 & 0.654 & 0.535 \\
   \cmidrule{2-10}
 & CHIMA-HM & 0.562 & 0.294 & 0.301 & 0.207 & 0.718 & 0.643 & 0.651 & 0.528 \\
 & CHIMA-SA & 0.567 & 0.303 & 0.307 & 0.213 & \bftab{0.722} & 0.660 & 0.666 & 0.536 \\
  \midrule 
\multirow{10}{*}{\rotatebox[origin=c]{90}{\begin{tabular}{c}
     Prachatai\\(1,250 articles)
\end{tabular}}} & Oracle & 0.853 & 0.836 & 0.839 & 0.754 & 1.000 & 1.000 & 1.000 & 1.000 \\
 & Lead-2 & 0.712 & 0.692 & 0.699 & 0.532 & 0.771 & 0.763 & 0.765 & 0.634 \\
 & Lead-3 & 0.694 & 0.675 & 0.680 & 0.539 & 0.762 & 0.754 & 0.755 & 0.638 \\
 & Lead-4 & 0.626 & 0.608 & 0.613 & 0.470 & 0.693 & 0.688 & 0.689 & 0.560 \\
 & HL & 0.381 & 0.235 & 0.286 & 0.086 & 0.354 & 0.215 & 0.264 & 0.071 \\
 \cmidrule{2-10}
 & HL-COS & 0.584 & 0.569 & 0.573 & 0.443 & 0.654 & 0.652 & 0.653 & 0.544 \\
 & BERTSUM & 0.753 & 0.719 & 0.726 & 0.616 & 0.839 & 0.817 & 0.822 & 0.758 \\
 & {}{GPTSUM} & {}{0.513} & {}{0.430} & {}{0.431} & {}{0.312} & {}{0.569} & {}{0.497} & {}{0.500} & {}{0.382} \\
 % & GPTSUM & 0.513 & 0.430 & 0.431 & 0.312 & 0.569 & 0.497 & 0.500 & 0.382 \\
 \cmidrule{2-10}
 & CHIMA-HM & \bftab{0.762} & \bftab{0.730} & \bftab{0.737} & \bftab{0.628} & \bftab{0.848} & 0.828 & 0.832 & \bftab{0.771} \\
 & CHIMA-SA & 0.759 & \bftab{0.730} & \bftab{0.737} & 0.627 & 0.846 & \bftab{0.829} & \bftab{0.833} & 0.770 \\
 \midrule
\multirow{10}{*}{\rotatebox[origin=c]{90}{\begin{tabular}{c}
     Thai PBS\\(178 articles)
\end{tabular}}} & Oracle & 0.683 & 0.585 & 0.557 & 0.486 & 1.000 & 1.000 & 1.000 & 1.000 \\
 & Lead-2 & 0.565 & 0.428 & 0.438 & 0.337 & 0.662 & 0.578 & 0.590 & 0.469 \\
 & Lead-3 & 0.568 & 0.438 & 0.442 & 0.339 & 0.686 & 0.607 & 0.613 & 0.511 \\
 & Lead-4 & 0.531 & 0.425 & 0.425 & 0.313 & 0.660 & 0.603 & 0.603 & 0.494 \\
 & HL & 0.446 & 0.339 & 0.380 & 0.124 & 0.342 & 0.221 & 0.252 & 0.056 \\
 \cmidrule{2-10}
 & HL-COS & 0.394 & 0.333 & 0.327 & 0.210 & 0.519 & 0.513 & 0.517 & 0.367 \\
 & BERTSUM & 0.578 & 0.433 & 0.439 & 0.344 & 0.690 & 0.611 & 0.620 & 0.504 \\
 & {}{GPTSUM} & {}{0.460} & {}{0.356} & {}{0.353} & {}{0.243} & {}{0.592} & {}{0.538} & {}{0.542} & {}{0.420} \\
 % & GPTSUM & 0.460 & 0.356 & 0.353 & 0.243 & 0.592 & 0.538 & 0.542 & 0.420 \\
 \cmidrule{2-10}
 & CHIMA-HM & 0.585 & 0.452 & 0.451 & 0.357 & 0.707 & 0.638 & 0.645 & 0.537 \\
 & CHIMA-SA & \bftab{0.587} & \bftab{0.462} & \bftab{0.461} & \bftab{0.366} & \bftab{0.716} & \bftab{0.654} & \bftab{0.660} & \bftab{0.561} \\
 \bottomrule
\end{tabular}
} % control the size of the table
\label{tab:result_rouge_bleu_individual}
\end{table*}

\begin{table*}[]
\centering
\caption{Model evaluation of binary classification outcomes at the document and sentence levels across all news sources.}
\resizebox{0.725\textwidth}{!}{ % control the size of the table
\begin{tabular}{llcccccc}
\toprule
 &  & \multicolumn{3}{c}{Document Level} & \multicolumn{3}{c}{Sentence Level} \\
\cmidrule(lr){3-5}
\cmidrule(l){6-8}
 &  & \multicolumn{1}{c}{Precision} & \multicolumn{1}{c}{Recall} & \multicolumn{1}{c}{F1} & \multicolumn{1}{c}{Precision} & \multicolumn{1}{c}{Recall} & \multicolumn{1}{c}{F1} \\
 \midrule
\multirow{8}{*}{\rotatebox[origin=c]{90}{\begin{tabular}{c} All {}{news sources}\\{}{(5,000 articles)} \end{tabular}}} 
% \multirow{8}{*}{\rotatebox[origin=c]{90}{All}} 
 & Lead-2 & 0.718 & 0.674 & 0.654 & 0.718 & 0.576 & 0.639 \\
 & Lead-3 & 0.628 & 0.823 & 0.671 & 0.628 & 0.756 & 0.686 \\
 & Lead-4 & 0.538 & 0.897 & 0.635 & 0.539 & 0.857 & 0.661 \\
 \cmidrule{2-8}
 & HL-COS & 0.549 & 0.976 & 0.652 & 0.440 & 0.977 & 0.607 \\
 & BERTSUM & 0.803 & 0.740 & 0.745 & 0.835 & 0.710 & 0.767 \\
 & {}{GPTSUM} & {}{0.366} & {}{0.678} & {}{0.442} & {}{0.343} & {}{0.606} & {}{0.438} \\
 % & GPTSUM & 0.366 & 0.678 & 0.442 & 0.343 & 0.606 & 0.438 \\
 \cmidrule{2-8}
 & CHIMA-HM & 0.796 & 0.793 & 0.771 & 0.791 & 0.768 & 0.779 \\
 & CHIMA-SA & 0.790 & 0.812 & \bftab{0.777} & 0.772 & 0.789 & \bftab{0.780} \\
 \bottomrule
\end{tabular}
} % control the size of the table
\label{tab:result_binary_all}
\end{table*}

\begin{table*}[]
\centering
\caption{Model evaluation of binary classification outcomes at the document and sentence levels across individual news sources.}
\resizebox{0.725\textwidth}{!}{ % control the size of the table
\begin{tabular}{llcccccc}
\toprule
 &  & \multicolumn{3}{c}{Document Level} & \multicolumn{3}{c}{Sentence Level} \\
 \cmidrule(lr){3-5}
\cmidrule(l){6-8}
 &  & \multicolumn{1}{c}{Precision} & \multicolumn{1}{c}{Recall} & \multicolumn{1}{c}{F1} & \multicolumn{1}{c}{Precision} & \multicolumn{1}{c}{Recall} & \multicolumn{1}{c}{F1} \\ 
 \midrule
\multirow{8}{*}{\rotatebox[origin=c]{90}
{\begin{tabular}{c}
     Thairath\\(3,436 articles)
\end{tabular}}} 
 & Lead-2 & 0.698 & 0.648 & 0.631 & 0.698 & 0.555 & 0.618 \\
 & Lead-3 & 0.623 & 0.808 & 0.662 & 0.623 & 0.742 & 0.677 \\
 & Lead-4 & 0.538 & 0.888 & 0.633 & 0.538 & 0.849 & 0.659 \\
 \cmidrule{2-8}
 & HL-COS & 0.562 & 0.972 & 0.661 & 0.450 & 0.971 & 0.615 \\
 & BERTSUM & 0.800 & 0.743 & 0.746 & 0.837 & 0.719 & 0.774 \\
 & {}{GPTSUM} & {}{0.364} & {}{0.682} & {}{0.442} & {}{0.343} & {}{0.608} & {}{0.438} \\
 % & GPTSUM & 0.364 & 0.682 & 0.442 & 0.343 & 0.608 & 0.438 \\
  \cmidrule{2-8}
 & CHIMA-HM & 0.797 & 0.789 & 0.771 & 0.795 & 0.768 & \bftab{0.781} \\
 & CHIMA-SA & 0.791 & 0.805 & \bftab{0.775} & 0.778 & 0.785 & \bftab{0.781} \\
 \midrule
\multirow{8}{*}{\rotatebox[origin=c]{90}{\begin{tabular}{c}
     The Standard\\(136 articles)
\end{tabular}}}
 & Lead-2 & 0.452 & 0.263 & 0.307 & 0.452 & 0.233 & 0.307 \\
 & Lead-3 & 0.502 & 0.439 & 0.432 & 0.502 & 0.388 & 0.438 \\
 & Lead-4 & 0.526 & 0.617 & 0.525 & 0.525 & 0.535 & 0.530 \\
  \cmidrule{2-8}
 & HL-COS & 0.539 & 0.994 & \bftab{0.669} & 0.539 & 0.992 & \bftab{0.699} \\
 & BERTSUM & 0.620 & 0.441 & 0.460 & 0.629 & 0.405 & 0.493 \\
 & {}{GPTSUM} & {}{0.557} & {}{0.654} & {}{0.554} & {}{0.553} & {}{0.590} & {}{0.571} \\
 % & GPTSUM & 0.557 & 0.654 & 0.554 & 0.553 & 0.590 & 0.571 \\
  \cmidrule{2-8}
 & CHIMA-HM & 0.598 & 0.573 & 0.531 & 0.609 & 0.543 & 0.574 \\
 & CHIMA-SA & 0.607 & 0.636 & 0.554 & 0.602 & 0.601 & 0.602 \\
 \midrule
\multirow{8}{*}{\rotatebox[origin=c]{90}{\begin{tabular}{c}
     Prachatai\\(1,250 articles)
\end{tabular}}}
 & Lead-2 & 0.826 & 0.811 & 0.774 & 0.826 & 0.716 & 0.767 \\
 & Lead-3 & 0.677 & 0.927 & 0.741 & 0.677 & 0.880 & 0.765 \\
 & Lead-4 & 0.553 & 0.969 & 0.669 & 0.555 & 0.951 & 0.701 \\
  \cmidrule{2-8}
 & HL-COS & 0.548 & 0.988 & 0.651 & 0.424 & 0.990 & 0.594 \\
 & BERTSUM & 0.853 & 0.797 & 0.803 & 0.880 & 0.771 & 0.822 \\
 & {}{GPTSUM} & {}{0.349} & {}{0.664} & {}{0.426} & {}{0.323} & {}{0.595} & {}{0.418} \\
 % & GPTSUM & 0.349 & 0.664 & 0.426 & 0.323 & 0.595 & 0.418 \\
  \cmidrule{2-8}
 & CHIMA-HM & 0.836 & 0.853 & 0.824 & 0.830 & 0.837 & \bftab{0.834} \\
 & CHIMA-SA & 0.829 & 0.873 & \bftab{0.830} & 0.805 & 0.859 & 0.831 \\
 \midrule
\multirow{8}{*}{\rotatebox[origin=c]{90}{\begin{tabular}{c}
     Thai PBS\\(178 articles)
\end{tabular}}}
 & Lead-2 & 0.539 & 0.528 & 0.503 & 0.539 & 0.479 & 0.507 \\
 & Lead-3 & 0.491 & 0.679 & 0.541 & 0.491 & 0.653 & 0.560 \\
 & Lead-4 & 0.437 & 0.785 & 0.535 & 0.439 & 0.771 & 0.559 \\
  \cmidrule{2-8}
 & HL-COS & 0.326 & 0.969 & 0.472 & 0.313 & 0.988 & 0.475 \\
 & BERTSUM & 0.655 & 0.515 & 0.541 & 0.645 & 0.461 & 0.538 \\
 & {}{GPTSUM} & {}{0.363} & {}{0.723} & {}{0.456} & {}{0.345} & {}{0.673} & {}{0.456} \\
 % & GPTSUM & 0.363 & 0.723 & 0.456 & 0.345 & 0.673 & 0.456 \\
  \cmidrule{2-8}
 & CHIMA-HM & 0.631 & 0.609 & 0.581 & 0.614 & 0.566 & 0.589 \\
 & CHIMA-SA & 0.631 & 0.658 & \bftab{0.609} & 0.612 & 0.628 & \bftab{0.620} \\
 \bottomrule
\end{tabular}
} % control the size of the table
\label{tab:result_binary_individual}
\end{table*}

Precision, recall, and F1 scores for classification tasks can be interpreted into different meaningful perspectives in text summarization. We calculate precision to measure the correctness of the predicted summary sentences by determining the proportion of correctly predicted sentences out of all predicted sentences:
\begin{equation}
\textrm{Precision} = \frac{\textrm{\# correctly predicted sentences}}{\textrm{\# predicted sentences}}.
\end{equation}

Recall is utilized to assess the coverage of the predicted sentences with the target sentences, measuring the proportion of correctly predicted sentences among all target sentences:
\begin{equation}
\textrm{Recall} = \frac{\textrm{\# correctly predicted sentences}}{\textrm{\# target sentences}}.
\end{equation}

Considering the imbalanced class distribution in extractive text summarization, the F1 score offers a more appropriate evaluation metric. 
The F1 score as a harmonic mean of precision and recall, effectively balances the trade-off between these two metrics, making it well-suited for imbalanced datasets.
\begin{equation}
\textrm{F1 score} = \frac{2 \times \textrm{Precision} \times \textrm{Recall}}{\textrm{Precision} + \textrm{Recall}}
\end{equation}

\subsubsection{Baselines}
We compared our proposed model with the following baselines:
\begin{itemize}
    \item \textbf{Oracle}\cite{summarunner}: The algorithm adopts a greedy strategy where predicted sentences are chosen based on their ROUGE scores.
    \item \textbf{Lead-$n$}: This approach utilizes a heuristic rule where the first $n$ sentences of the article are selected as the predicted summary. For example, \textbf{Lead-2} selects the first two sentences, \textbf{Lead-3} selects the first three, and \textbf{Lead-4} selects the first four sentences.
    \item \textbf{HL}: This method leverages the headline as a predicted summary.
    \item \textbf{HL-COS}: This model utilizes the cosine similarity between the headline's representation and each sentence's representation in the article's body as a predicted score.
    \item {\textbf{BERTSUM}\cite{bertsum}: This model is a variant of BERT tailored to the extractive text summarization task by stacking a summarization layer on top of the BERT encoder.
    \item {{}{\textbf{GPTSUM} \mbox{\cite{chatgptsum}}: This model utilizes ChatGPT's capabilities to generate an extractive summary.}}}
\end{itemize}

\subsection{Results} 
In this section, we present two aspects of the evaluation results: the performance of summary sentence selection and the binary classification outcomes.
For each aspect, the evaluation results are reported in two tables: one showing the results across all news sources and the other showing the results for each news source individually.
Specifically, the performances of summary sentence selection are reported in Table \ref{tab:result_rouge_bleu_all} and \ref{tab:result_rouge_bleu_individual}, while the binary classification performances are reported in Table \ref{tab:result_binary_all} and \ref{tab:result_binary_individual}.

\begin{figure}[t!]
\centering
\includegraphics[width = 0.48\textwidth]{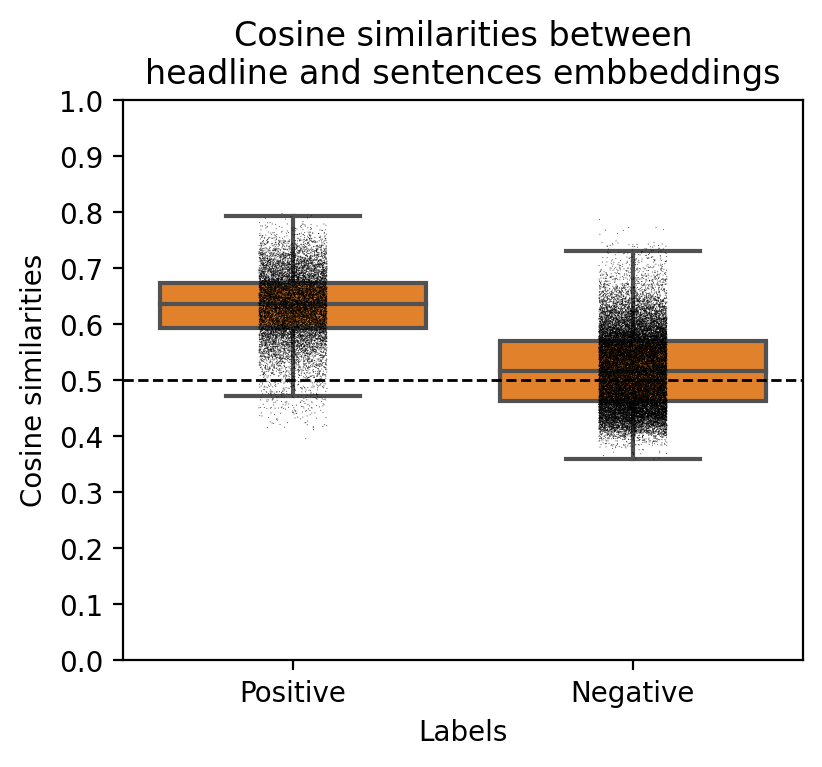}
\caption{Headline-body cosine similarities for summary and non-summary sentences.}
\label{fig:fig4_pos_neg_cos_only}
\end{figure}

\begin{figure}[t!]
\centering
\includegraphics[width = 0.48\textwidth]{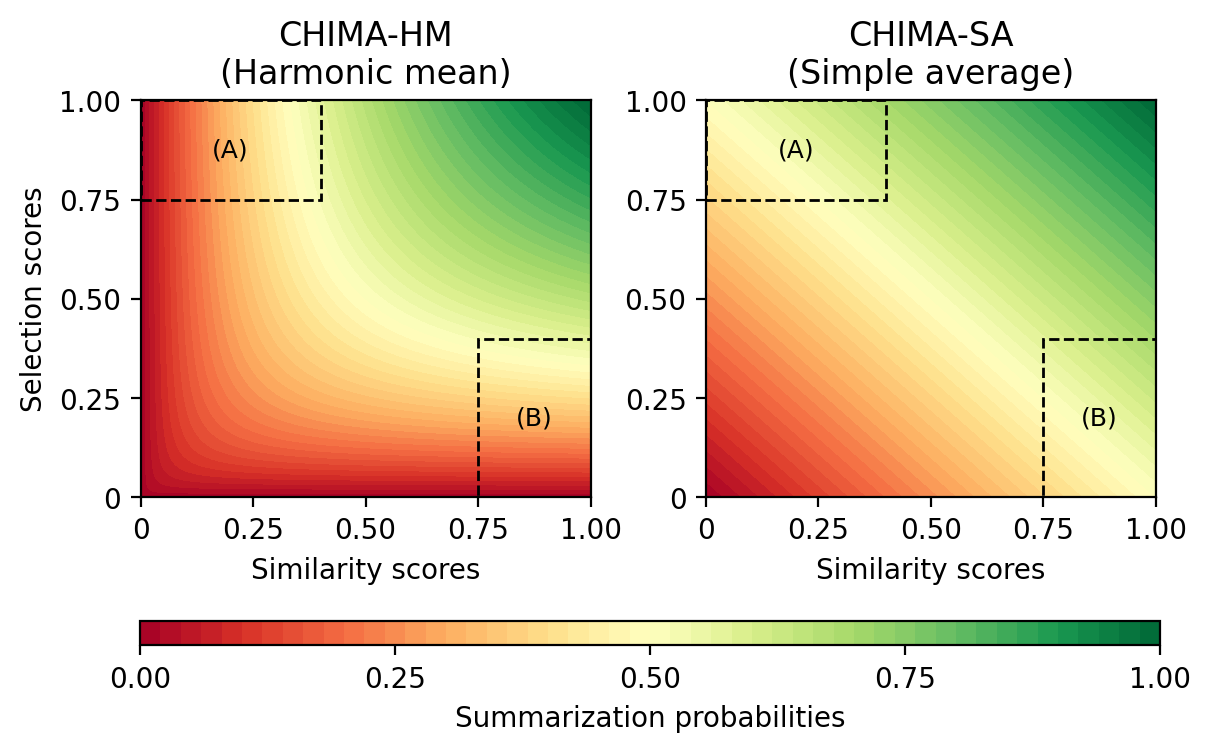}
\caption{Contour plots of summarization probabilities based on varying
selection and similarity scores.}
\label{fig:fig5_compare_hm_sa}
\end{figure}

\begin{figure}[t!]
\centering
\includegraphics[width = 0.4625\textwidth]{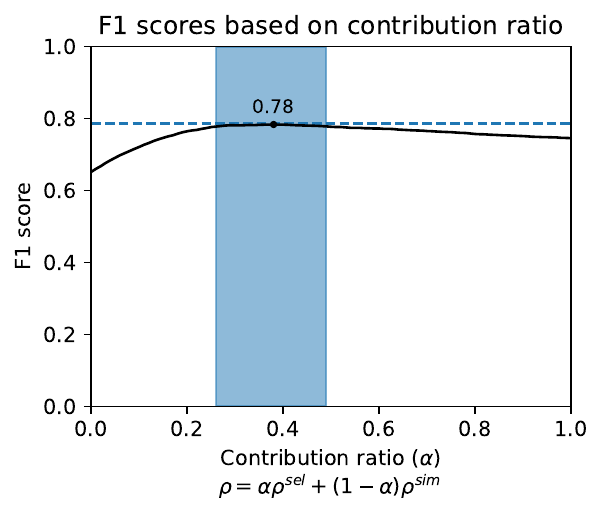}
\caption{F1 scores with varying contribution ratios in the aggregation of the selection and semantic similarity scores.}
\label{fig:fig6_f1_contribution_ratio}
\end{figure}

\begin{figure*}[t!]
\centering
\includegraphics[width = 0.68\textwidth]{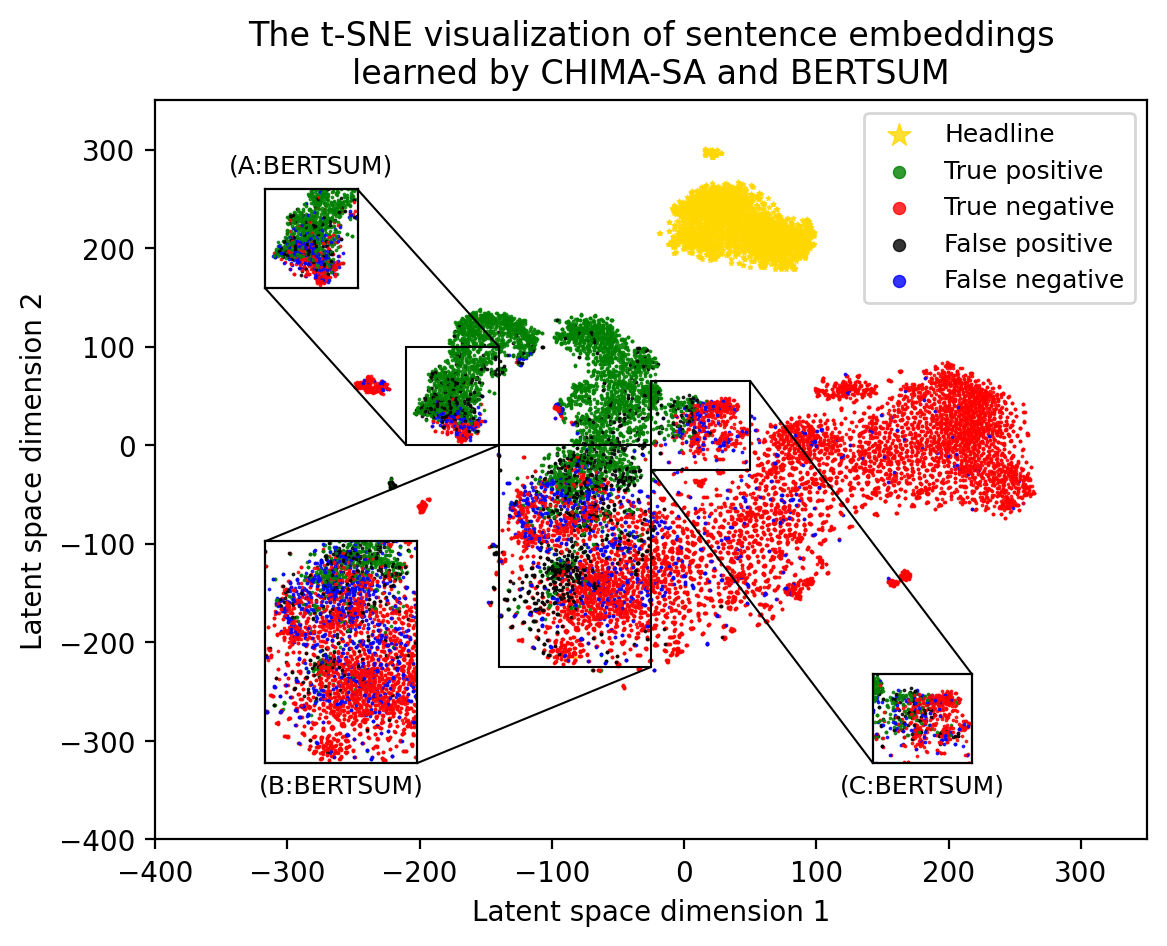}
\caption{ {}{ Visualization of sentence embeddings learned by CHIMA-SA and BERTSUM using t-SNE. The main region displays embeddings from CHIMA-SA, while inset plots of regions A, B, and C, correspond to embeddings from BERTSUM.} Each color represents a different class: true positives (green), true negatives (red), false positives (black), false negatives (blue), and headline embeddings (yellow).}
\label{fig:fig7_tsne}
\end{figure*}

\subsubsection{Evaluation of the extracted summary using $n$-gram overlap metrics} 
We first evaluated the summary results extracted by each model using ROUGE and BLEU scores.
The predicted summary is compared against abstractive and extractive summaries.
Particularly,  abstractive summaries are provided in the dataset, whereas extractive summaries are generated by Oracle.  
Table \ref{tab:result_rouge_bleu_all} summarizes the performance of all models on the entire test set. 
Notably, Oracle served as the upper limit for all metrics, as it extracted the summary sentences using a greedy approach optimized from the provided abstractive summary. 

Among the heuristic baselines, Lead-2 performed relatively well, capturing more than half of the target summary (ROUGE $>$ 0.5). By including an additional sentence, Lead-3 slightly improved upon Lead-2. 
In contrast, Lead-4 demonstrated inferior performance compared to both Lead-2 and Lead-3. 
The drop in performance of Lead-4 highlights the negative impact of including irrelevant sentences in the summary. 
These findings suggest that significant information in articles often appears near the beginning, aligning with the statistics shown in Figure \ref{fig:fig3_dataset_analysis}(d).

HL, which used only the headline as a summary, resulted in the poorest performance due to the limited information typically contained in the headline. 
Meanwhile, HL-COS outperformed HL, suggesting the advantages of semantically integrating headlines as guidance in the summary sentence selection process. 
By incorporating semantic similarity between the headline and body sentences, this approach enables the summary to expand beyond the headline while maintaining a focus on the article's key points.

BERTSUM outperformed all heuristic baselines, showcasing the benefits of adopting a pre-trained language model.
{}{Although GPTSUM demonstrates greater complexity than other models, it achieved the second-lowest summarization performance. The primary issue may stem from the challenges associated with unsupervised learning, where these models are not explicitly trained for specific tasks but instead rely on general-purpose training. Additionally, this result highlights the limitations of current multilingual LLMs, such as ChatGPT, when applied to specific downstream tasks in low-resource languages like Thai.}
However, our proposed CHIMA-SA and CHIMA-HM consistently surpassed BERTSUM {}{and GPTSUM} across all metrics. 
While CHIMA-SA slightly underperformed CHIMA-HM on ROUGE-1 when evaluated on the abstractive summary, it achieved the highest scores on all other metrics. 
These results demonstrate that exploiting both selection and similarity scores guided by the headline at the embedding level can effectively enhance sentence extraction for summarization.

Table \ref{tab:result_rouge_bleu_individual} shows that CHIMA-HM and CHIMA-SA outperformed other models on the ``Thairath'', ``Prachatai'', and ``Thai PBS datasets'', while HL-COS performed competitively on "The Standard" dataset. 
The stronger performance of HL-COS may be attributed to the unique writing style of each news source. 
Nevertheless, the outperformance of CHIMA-HM and CHIMA-SA across various datasets demonstrates the robustness and generalization of our models in handling diverse writing styles across different news sources.

\begin{figure*}[t!]
\centering
\includegraphics[width = 0.8\textwidth]{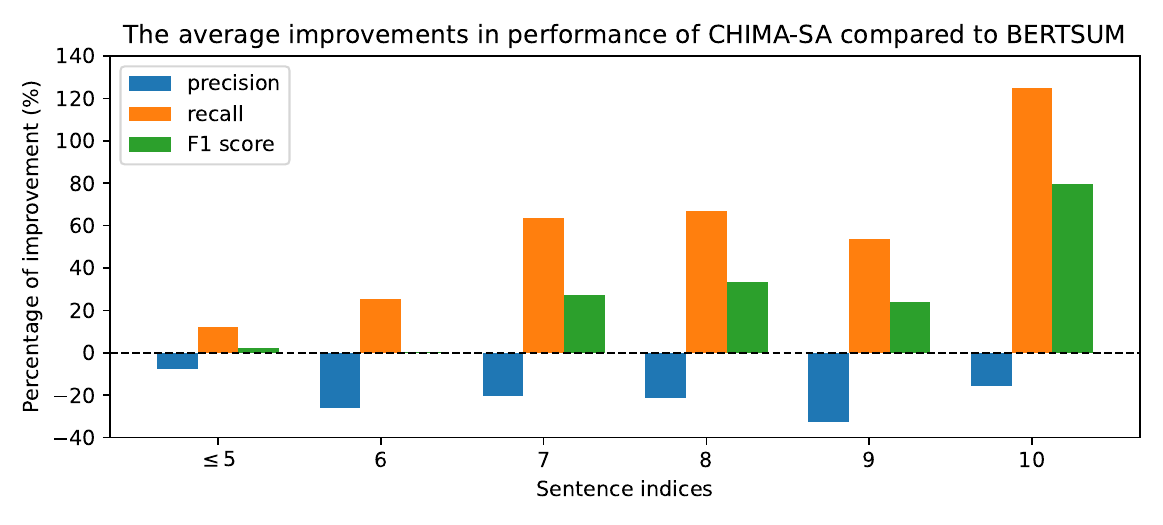}
\caption{Percentage improvements in precision, recall, and F1 scores across sentence indices.}
\label{fig:fig8_score_improvement}
\end{figure*}

\subsubsection{Assessment of the summary sentence selection as a binary classification task using precision, recall, and F1 scores}

To further evaluate the summarization performance, we formalized the extracted summary as a binary classification task.
The evaluation is performed at the document and sentence levels using precision, recall, and F1 scores.
The target reference is the extractive summary determined by the Oracle method.
In the document-level evaluation, these metrics were calculated independently for each article and then averaged across all articles to provide the model's overall performance. 
On the other hand, the sentence-level assessment combined all sentences from all articles and computed the metrics collectively. 
The document-level evaluation assesses the model's ability to output cohesive summaries that capture the overall essence of each document.
In contrast, the sentence-level evaluation focuses on the precision and relevance of individual sentences. 
Together, these assessments offer a comprehensive understanding of the model's performance in identifying and summarizing key information.

According to Table \ref{tab:result_binary_all}, all baseline models demonstrated strong performance in either precision or recall at the document and sentence levels. 
HL-COS achieved high recall but lower precision, while BERTSUM excelled in precision but had lower recall. 
{}{Despite exhibiting higher recall than precision, GPTSUM showed poor performance in both metrics.}
Although these models performed well in specific metrics, they fell short of CHIMA in terms of the F1 score.
In contrast, CHIMA-SA and CHIMA-HM delivered both high and well-balanced performance across precision and recall, resulting in the best F1 score. 
Table \ref{tab:result_binary_individual} additionally highlights their superior F1 performance on most datasets i.e. ``Thairath'', ``Prachatai'', and ``Thai PBS'', while HL-COS achieved promising results on the ``The Standard'' dataset.

These results emphasize the consistent effectiveness of integrating headline information in CHIMA models, enhancing their ability to recall key sentences and output more informative summaries of Thai news articles.

\section{Discussions}\label{sec:discuss}
The results indicate that incorporating a headline as additional contextual information in extractive summarization models enhances the performance of sentence selection. In this section, we discuss three main aspects of this improvement: the impact of utilizing headlines to guide the model for summary sentence selection, a model performance analysis, and an in-depth case study.

\subsection{Impact of utilizing headline for model guidance in summary sentence selection}
\subsubsection{Headline-body semantic similarity}
To investigate the impact of the headline on individual article's body sentences, we computed cosine similarity to measure the semantic similarities between the article's body sentence embeddings and their corresponding headline embeddings.
Then, we compared them against positive and negative labels.

Most of the cosine similarities were above 0.5, indicating a high likelihood of these sentences being selected as summary sentences (see Figure \ref{fig:fig4_pos_neg_cos_only}).
Although the distinction between the positive and negative labels was subtle, the positive label class had a slightly but significantly higher median cosine similarity compared to the negative label class (p-value < 0.001 for unpaired t-test). 
Specifically, the medians of cosine similarity for the positive and negative classes were 0.64 and 0.52, respectively. 
This suggests that the headline embeddings align more closely with the summary sentences than non-summary sentences. 
Nevertheless, relying solely on this cosine similarity as the predicted probability in the HL-COS model resulted in unsatisfactory performance across most metrics, except for recall, as mentioned earlier.
In contrast, CHIMA models, which integrated headline similarity and sentence selection scores demonstrated performance improvements in most metrics. 
These results support the benefit of using the headline semantic similarity as part of the sentence selection step rather than as a standalone model.

\begin{figure*}[t!]
\centering
\includegraphics[width = 0.96\textwidth]{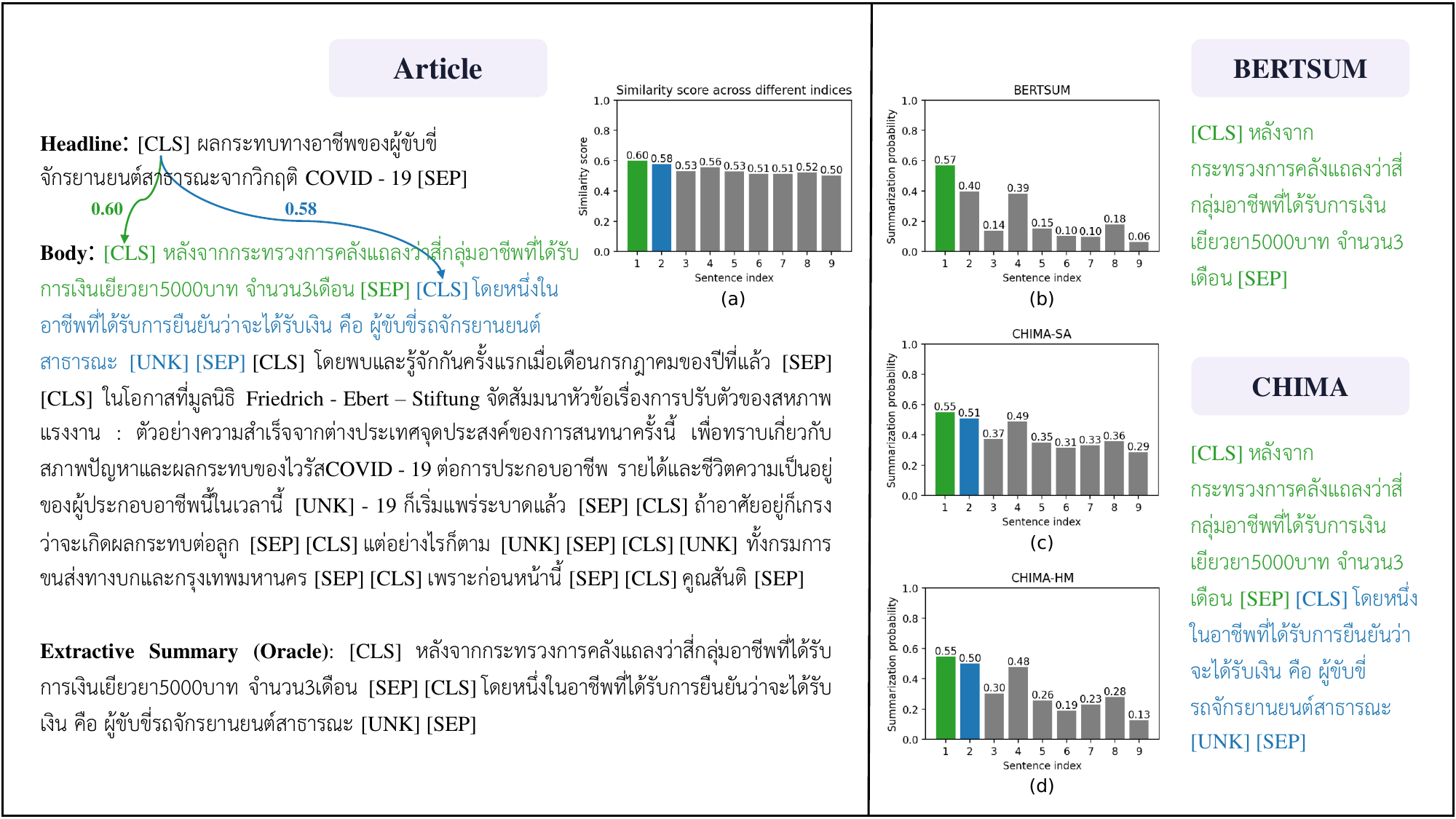}
\caption{A case study demonstrating the benefit of considering the headline when predicting an extractive summary.}
\label{fig:fig9_case_study}
\end{figure*}

\subsubsection{Comparative mechanisms of headline-guided aggregation functions}
When comparing the experimental results among CHIMA models, the slightly lower performance of CHIMA-HM may be influenced by the stricter conditions imposed by the harmonic mean.
According to its formula, the harmonic mean requires both of the body sentence selection score ($\rho^{sel}$) and the semantic similarity between the headline and body sentence ($\rho^{sim}$) to be high to constitute a high final summarization probability ($\rho$). 
In contrast, CHIMA-SA selects the sentence as a summary if either $\rho^{sel}$ or $\rho^{sim}$ is high as it utilizes a simple average function to combine these scores.

In Figure \ref{fig:fig5_compare_hm_sa}, we compared two aggregation functions: the harmonic mean in CHIMA-HM and the simple average in CHIMA-SA.
Each point in the contour plot represents the output probability of a sentence being included in the summary. 
We now consider two regions with high predictive scores in either selection score or similarity score. 
Region (A) represents areas with high selection scores but low headline similarity scores, whereas Region (B) shows the reverse — low selection scores but high headline similarity scores. 
In the CHIMA-HM plot, these regions yield lower summarization probabilities, reflected by red shading, due to the harmonic mean’s strict requirement. This criterion limits to select only sentences with strong confidence in both scores. 
In contrast, in the CHIMA-SA plot, regions (A) and (B) have higher summarization probabilities, shown in more yellow-green shading, because the simple average allows sentences to achieve a high summarization probability if either the selection or similarity score is high.
This flexibility enables CHIMA-SA to select sentences even when only one score is elevated, potentially producing extensive summarization.

\subsubsection{Balancing selection and similarity scores for optimal extractive summarization}
Our CHIMA models adopt the concept of weighted average to combine the selection score and the semantic similarity score to achieve a higher F1 score compared to relying on either score alone. In this section, we further investigate the effects of headline integration in enhancing the predictive summarization probability.  
In particular, we analyze the impact of the contribution ratio ($\alpha$) on the F1 score. This variable controls the contribution of the selection score ($\rho^{sel}_i$) and the semantic similarity score ($\rho^{sim}_i$), as follows:

\begin{equation}
\rho_i = \mathbf{\alpha} \rho^{sel}_i + (1-\mathbf{\alpha}) \rho^{sim}_i.
\label{eq:alpha_formula}
\end{equation}

As shown in Figure \ref{fig:fig6_f1_contribution_ratio}, the F1 score rises as $\alpha$ increases, reaching its peak when $\alpha$ is between 0.25 and 0.50. Beyond this point, the F1 score gradually declines. 
These results suggest that incorporating the headline with an appropriately weighted combination of the selection and semantic similarity scores achieves superior performance compared to models relying on either score individually.
Furthermore, the best result at $\alpha$ around 0.4 indicates that both scores are essential for the summarization task. The slightly higher contribution of semantic similarity highlights the important role of the headline in helping the model select summary sentences effectively.

\subsection{Analysis of model performance across sentence embeddings and their positions within articles}
\subsubsection{Headline-guided separation of embedding space and classification outcomes}

Although BERTSUM relies on the same backbone model for sentence representations, our proposed CHIMA-SA achieves better separation of classification outcomes in the sentence embedding space as visualized using t-SNE \cite{tsne} in Figure \ref{fig:fig7_tsne}. 
As expected, summary sentences (green and black) are more tightly clustered and closer to the isolated headlines (yellow), whereas non-summary sentences (red and blue) are more dispersed.
The separation between true positives and true negatives is fairly distinct, indicating that the model has learned to effectively differentiate between positive and negative classes.

Still, some degree of overlap between them remains evident, as illustrated in Figure \ref{fig:fig7_tsne} A, B, and C. 
These areas contain false positives (black) and false negatives (blue), reflecting misclassified sentences.
The overlapping areas are where classifiers face challenges in differentiating between summary and non-summary sentences. 
Under these circumstances, CHIMA-SA performed better in these areas by recalling more true positives, particularly in regions A and B, thanks to its use of headline similarity. 
In contrast, BERTSUM struggled more in these overlapping regions, with fewer true positives, as its predictions were more conservative. 
This evidence suggests that CHIMA-SA's integration of headline information helps it overcome ambiguity better than BERTSUM.
Our embedding analysis aligns with the findings reported in Table \ref{tab:result_binary_all}, in which BERTSUM demonstrates a higher precision, while CHIMA-SA achieves higher recall and F1 scores.

\subsubsection{Improvements of summary selection across sentence positions within article}
We analyze the model performance in selecting summary sentences across sentence indices within the article. For each index, average precision, recall, and F1 scores were calculated. We then compared CHIMA-SA and BERTSUM based on these metrics and computed the average percentage improvements in these metrics across the first 10 sentences (see Figure \ref{fig:fig8_score_improvement}). 

CHIMA-SA performed on par with BERTSUM in the initial sentences (first five) of the article, while prioritizing improved recall at the expense of some precision. 
Moreover, the performance of CHIMA-SA improved over BERTSUM for sentences appearing later in the news articles. 
Notably, CHIMA-SA achieved up to 130\% recall improvement and an 80\% F1 score gain at the $10^{\textrm{th}}$ sentence.
While precisions decrease slightly, the significant gains in recall and F1 scores highlight our model's ability to capture more relevant information without compromising overall performance. 
These improvements position CHIMA-SA as a compelling alternative to BERTSUM, particularly excelling at identifying relevant information despite being located in the middle or at the end of the article.

\subsection{Case study: Effective recall of headline-guided summary sentences}
Finally, we present a case study based on an article in the test set to demonstrate the effectiveness of the proposed CHIMA models in recalling key sentences.
Figure \ref{fig:fig9_case_study} shows its headline, article's body, and extractive summary produced by the Oracle.
Panel (a) shows the semantic similarity based on cosine similarity for each sentence index, where the top three most similar sentences are in the first, second, and fourth positions, respectively.

The summary obtained by BERTSUM included only the first sentence, while the predicted probability for the second sentence was 0.4, as shown in Panel (b). This may be attributed to BERTSUM's limitation of relying solely on the article's body to extract the summary.
On the other hand, CHIMA-SA and CHIMA-HM which incorporated the headline were able to recover the second sentence as another sentence in the summary, with the probability of CHIMA-SA and CHIMA-HM of 0.51 and 0.50, respectively.
This outcome demonstrates the benefit of considering the headline to recall important sentences in the extractive summary.

\section{Conclusion}\label{sec:conclusion}
In this work, we incorporate headline information as a guiding feature to enhance extractive summarization for Thai news articles. The experimental results demonstrate that the proposed CHIMA models consistently outperform baseline methods across multiple evaluation metrics, including ROUGE, BLEU, and F1 scores. This improved performance, observed at both the document and sentence levels, stems from the CHIMA models’ ability to recall key summary sentences more effectively than methods relying solely on the article's body. 

Among the proposed models, CHIMA-SA offers greater flexibility due to its simple averaging approach, which selects sentences with high scores in either the selection or similarity modules. In contrast, CHIMA-HM employs a harmonic mean approach that requires both scores to be consistently high. Further analysis of the contribution ratio ($\alpha$) confirms that both selection and semantic similarity scores play significant roles in determining the final summarization probability. 

Our analysis of sentence embedding using t-SNE visualization reveals CHIMA-SA’s effectiveness in distinguishing between sentence classes. In the latent space, sentences with close semantic proximity to headline embeddings are more effectively recalled, particularly in areas prone to confusion with non-summary sentences. Additionally, a case study highlights the effectiveness of our approach, showcasing improved coverage of extracted summaries.

By incorporating headline-body similarity as guidance, we have demonstrated that the proposed CHIMA models successfully recall key sentences scattered throughout news articles, resulting in more complete summaries. 
This integration of semantic similarity with selection mechanisms not only enhances summary quality but also lays a foundation for more adaptable and context-aware summarization approaches.

\bibliographystyle{IEEEtran}
\bibliography{ms}

% Generated by IEEEtran.bst, version: 1.14 (2015/08/26)
\begin{thebibliography}{10}
\providecommand{\url}[1]{#1}
\csname url@samestyle\endcsname
\providecommand{\newblock}{\relax}
\providecommand{\bibinfo}[2]{#2}
\providecommand{\BIBentrySTDinterwordspacing}{\spaceskip=0pt\relax}
\providecommand{\BIBentryALTinterwordstretchfactor}{4}
\providecommand{\BIBentryALTinterwordspacing}{\spaceskip=\fontdimen2\font plus
\BIBentryALTinterwordstretchfactor\fontdimen3\font minus \fontdimen4\font\relax}
\providecommand{\BIBforeignlanguage}[2]{{%
\expandafter\ifx\csname l@#1\endcsname\relax
\typeout{** WARNING: IEEEtran.bst: No hyphenation pattern has been}%
\typeout{** loaded for the language `#1'. Using the pattern for}%
\typeout{** the default language instead.}%
\else
\language=\csname l@#1\endcsname
\fi
#2}}
\providecommand{\BIBdecl}{\relax}
\BIBdecl

\bibitem{chatgptsum}
H.~Zhang, X.~Liu, and J.~Zhang, ``Extractive summarization via chatgpt for faithful summary generation,'' 2023.

\bibitem{bertsum}
Y.~Liu, ``Fine-tune bert for extractive summarization,'' 2019.

\bibitem{summareranker}
M.~Ravaut, S.~Joty, and N.~F. Chen, ``Summareranker: A multi-task mixture-of-experts re-ranking framework for abstractive summarization,'' 2023.

\bibitem{matchsum}
M.~Zhong, P.~Liu, Y.~Chen, D.~Wang, X.~Qiu, and X.~Huang, ``Extractive summarization as text matching,'' in \emph{Proceedings of the 58th Annual Meeting of the Association for Computational Linguistics}.\hskip 1em plus 0.5em minus 0.4em\relax Online: Association for Computational Linguistics, Jul. 2020, pp. 6197--6208.

\bibitem{sum_survey}
A.~P. Widyassari, S.~Rustad, G.~F. Shidik, E.~Noersasongko, A.~Syukur, A.~Affandy, and D.~R. I.~M. Setiadi, ``Review of automatic text summarization techniques \& methods,'' \emph{Journal of King Saud University - Computer and Information Sciences}, vol.~34, no.~4, pp. 1029--1046, 2022.

\bibitem{economic_new}
K.~Tawong, P.~Pholsukkarn, P.~Noawaroongroj, and T.~Siriborvornratanakul, ``Economic news using lstm and gru models for text summarization in deep learning,'' \emph{Journal of Data, Information and Management}, vol.~6, no.~1, pp. 29--39, 2024.

\bibitem{th_mf}
S.~Nathonghor and D.~Wichadakul, ``Extractive text summarization for thai travel news based on keyword scored in thai language,'' in \emph{Proceedings of the 2020 2nd International Conference on Information Technology and Computer Communications}, ser. ITCC '20.\hskip 1em plus 0.5em minus 0.4em\relax New York, NY, USA: Association for Computing Machinery, 2020, p. 32–36.

\bibitem{rel_work_th2}
P.~Porntrakoon, C.~Moemeng, and P.~Santiprabhob, ``Text summarization for thai food reviews using simplified sentiment analysis,'' in \emph{2021 18th International Joint Conference on Computer Science and Software Engineering (JCSSE)}, 2021, pp. 1--5.

\bibitem{lstm}
S.~Hochreiter and J.~Schmidhuber, ``Long short-term memory,'' \emph{Neural computation}, vol.~9, no.~8, pp. 1735--1780, 1997.

\bibitem{gru}
K.~Cho, B.~van Merri{\"e}nboer, C.~Gulcehre, D.~Bahdanau, F.~Bougares, H.~Schwenk, and Y.~Bengio, ``Learning phrase representations using {RNN} encoder{--}decoder for statistical machine translation,'' in \emph{Proceedings of the 2014 Conference on Empirical Methods in Natural Language Processing ({EMNLP})}, A.~Moschitti, B.~Pang, and W.~Daelemans, Eds.\hskip 1em plus 0.5em minus 0.4em\relax Doha, Qatar: Association for Computational Linguistics, Oct. 2014, pp. 1724--1734.

\bibitem{thaisum_github}
N.~Chumpolsathien, ``Using knowledge distillation from keyword extraction to improve the informativeness of neural cross-lingual summarization,'' Master's thesis, Beijing Institute of Technology, 2020.

\bibitem{bert}
J.~Devlin, M.-W. Chang, K.~Lee, and K.~Toutanova, ``{BERT}: Pre-training of deep bidirectional transformers for language understanding,'' in \emph{Proceedings of the 2019 Conference of the North {A}merican Chapter of the Association for Computational Linguistics: Human Language Technologies, Volume 1 (Long and Short Papers)}.\hskip 1em plus 0.5em minus 0.4em\relax Minneapolis, Minnesota: Association for Computational Linguistics, Jun. 2019, pp. 4171--4186.

\bibitem{transformer}
A.~Vaswani, N.~Shazeer, N.~Parmar, J.~Uszkoreit, L.~Jones, A.~N. Gomez, L.~u. Kaiser, and I.~Polosukhin, ``Attention is all you need,'' in \emph{Advances in Neural Information Processing Systems}, I.~Guyon, U.~V. Luxburg, S.~Bengio, H.~Wallach, R.~Fergus, S.~Vishwanathan, and R.~Garnett, Eds., vol.~30.\hskip 1em plus 0.5em minus 0.4em\relax Curran Associates, Inc., 2017.

\bibitem{presumm}
Y.~Liu and M.~Lapata, ``Text summarization with pretrained encoders,'' \emph{CoRR}, vol. abs/1908.08345, 2019.

\bibitem{sbert}
N.~Thakur, N.~Reimers, J.~Daxenberger, and I.~Gurevych, ``Augmented {SBERT}: Data augmentation method for improving bi-encoders for pairwise sentence scoring tasks,'' in \emph{Proceedings of the 2021 Conference of the North American Chapter of the Association for Computational Linguistics: Human Language Technologies}.\hskip 1em plus 0.5em minus 0.4em\relax Online: Association for Computational Linguistics, Jun. 2021, pp. 296--310.

\bibitem{th_fast}
K.~Jearanaitanakij, S.~Boonpong, K.~Teainnagrm, T.~Thonglor, T.~Kullawan, and C.~Yongpiyakul, ``Fast hybrid approach for thai news summarization,'' \emph{Engineering and Technology Horizons}, vol.~41, no.~3, p. 410307, Sep. 2024.

\bibitem{bart}
M.~Lewis, Y.~Liu, N.~Goyal, M.~Ghazvininejad, A.~Mohamed, O.~Levy, V.~Stoyanov, and L.~Zettlemoyer, ``{BART:} denoising sequence-to-sequence pre-training for natural language generation, translation, and comprehension,'' \emph{CoRR}, vol. abs/1910.13461, 2019.

\bibitem{th_key}
P.~Ngamcharoen, N.~Sanglerdsinlapachai, and P.~Vejjanugraha, ``Automatic thai text summarization using keyword-based abstractive method,'' in \emph{2022 17th International Joint Symposium on Artificial Intelligence and Natural Language Processing (iSAI-NLP)}, 2022, pp. 1--5.

\bibitem{summarunner}
R.~Nallapati, F.~Zhai, and B.~Zhou, ``Summarunner: {A} recurrent neural network based sequence model for extractive summarization of documents,'' \emph{CoRR}, vol. abs/1611.04230, 2016.

\bibitem{residual}
K.~He, X.~Zhang, S.~Ren, and J.~Sun, ``Deep residual learning for image recognition,'' in \emph{2016 IEEE Conference on Computer Vision and Pattern Recognition (CVPR)}, 2016, pp. 770--778.

\bibitem{layer_norm}
J.~L. Ba, J.~R. Kiros, and G.~E. Hinton, ``Layer normalization,'' 2016.

\bibitem{pythainlp}
W.~Phatthiyaphaibun, K.~Chaovavanich, C.~Polpanumas, A.~Suriyawongkul, L.~Lowphansirikul, and P.~Chormai, ``{P}y{T}hai{NLP}: {T}hai natural language processing in {P}ython,'' Jun. 2016.

\bibitem{ROUGE}
C.-Y. Lin, ``{ROUGE}: A package for automatic evaluation of summaries,'' in \emph{Text Summarization Branches Out}.\hskip 1em plus 0.5em minus 0.4em\relax Barcelona, Spain: Association for Computational Linguistics, Jul. 2004, pp. 74--81.

\bibitem{tsne}
L.~van~der Maaten and G.~Hinton, ``Visualizing data using t-sne,'' \emph{Journal of Machine Learning Research}, vol.~9, no.~86, pp. 2579--2605, 2008.

\end{thebibliography}

\end{document}